%% file: neurips_2026.tex
\documentclass{article}
\usepackage[main, final]{neurips_2026}
\usepackage{listings}
\usepackage{tcolorbox}
\usepackage{xcolor}
\usepackage{xfp}
\usepackage{nicematrix}
\usepackage{enumitem}
\usepackage{pifont}
\usepackage{microtype}
\usepackage{graphicx}
\usepackage{subcaption}
\usepackage{booktabs} 
\usepackage{fontawesome5}
\usepackage[colorlinks]{hyperref}

\usepackage[utf8]{inputenc} 
\usepackage[T1]{fontenc}    
\usepackage{hyperref}      
\usepackage{url}           
\usepackage{booktabs}      
\usepackage{amsfonts}      
\usepackage{nicefrac}      
\usepackage{microtype}      
\usepackage{xcolor}

\usepackage{amsmath, amssymb, amsthm}
\usepackage{algorithm}
\usepackage{algorithmic}
\usepackage{geometry}
\usepackage{amsmath}
\usepackage{amssymb}
\usepackage{mathtools}
\usepackage{amsthm}

\usepackage{listings}

\lstdefinestyle{promptstyle}{
  basicstyle=\ttfamily\scriptsize,
  breaklines=true,
  breakatwhitespace=false,
  columns=fullflexible,
  keepspaces=true,
  showstringspaces=false
}

\usepackage[capitalize,noabbrev]{cleveref}

\theoremstyle{plain}

\theoremstyle{definition}

\theoremstyle{remark}

\tcbuselibrary{breakable, skins}
\newtcolorbox{planbox}[1]{
  enhanced,
  colback=white,
  colframe=logoCyan!80,
  coltitle=logoBlue,
  fonttitle=\bfseries\sffamily,
  title=#1,
  titlerule=0.8pt,
  boxrule=1pt,
  left=1mm, right=1mm, top=1mm, bottom=1mm,
  boxsep=1mm,
  width=1.1\textwidth,
  before upper=\smallskip,
}

\newcommand{\tightcolorbox}[2]{%
  \begingroup
  \setlength{\fboxsep}{1pt}%
  \colorbox{#1}{#2}%
  \endgroup
}

\usepackage{ragged2e}
\newtcolorbox{examplebox}{
  colback=green!10,
  colframe=black!15,
  boxrule=0.3pt,
  arc=1pt,
  left=3pt,
  right=3pt,
  top=3pt,
  bottom=3pt,
  fontupper=\footnotesize,
  before upper={\justifying}
}

\usepackage{ragged2e}
\newtcolorbox{examplebox2}{
  colback=red!10,
  colframe=black!15,
  boxrule=0.3pt,
  arc=1pt,
  left=3pt,
  right=3pt,
  top=3pt,
  bottom=3pt,
  fontupper=\footnotesize,
  before upper={\justifying}
}

\definecolor{logoBlue}{HTML}{1A4E8A}
\definecolor{logoCyan}{HTML}{56BBCC}

\definecolor{IllinoisDarkerBlue}{HTML}{13294B}
\definecolor{IllinoisBlue}{HTML}{3C78D8}
\definecolor{IllinoisOrange}{HTML}{FF5F05}
\definecolor{ForestGreen}{RGB}{34,139,34}
\definecolor{CustomPurple}{HTML}{884AB2}

\newcommand{\modelname}{\textbf{\textcolor{IllinoisOrange}{VisAnom}\textcolor{IllinoisBlue}{Reasoner}}}

\newcommand{\dataset}{\textbf{\textsc{\textcolor{IllinoisOrange}{VisAnom}\textcolor{IllinoisDarkerBlue}{Bench}}}}

\newcommand{\modelnamenc}{{VisAnomReasoner}}
\newcommand{\datasetnc}{{\textsc{VisAnomBench}}}

\newcommand{\uparr}{\textcolor{ForestGreen}{$\uparrow$}}
\newcommand{\downarr}{\textcolor{red}{$\downarrow$}}

\newcommand{\downarrg}{\textcolor{ForestGreen}{$\downarrow$}}
\usepackage{tikz}
\usepackage{pgfplots}
\usepackage{pgf-pie}
\usepackage{wrapfig}
\pgfplotsset{compat=1.18} 

\usepackage{xspace}
\newcommand*{\eg}{\emph{e.g.},\@\xspace}

\newcommand{\comp}[2]{%
  \edef\diffval{\fpeval{round(abs((#1)-(#2)),2)}}%
  \ifdim #1pt>#2pt
    \,\uparr\,\textcolor{ForestGreen}{\diffval}%
  \else\ifdim #1pt<#2pt
    \,\downarr\,\textcolor{red}{\diffval}%
  \else
    \,\textcolor{gray!70}{0.00}%
  \fi\fi
}

\definecolor{genVLM}{RGB}{212,212,212}       

\definecolor{smVLM}{RGB}{235,235,235}        

\definecolor{FM}{RGB}{255,245,210}           

\definecolor{detector}{RGB}{232,244,255}

\definecolor{specLM}{RGB}{245,235,255}     

\definecolor{ourVLM}{named}{IllinoisOrange}

\usepackage{wrapfig}

\usepackage[textsize=tiny]{todonotes}

\title{Tiny but Trusted: Efficient Vision-Language Reasoning for Time-Series Anomaly Detection}

\author{
Xiaona Zhou$^{1}$,
Muntasir Wahed$^{1}$,
Tianjiao Yu$^{1}$,
Constantin Brif$^{2}$,
Ismini Lourentzou$^{1}$ \\
$^{1}$University of Illinois Urbana-Champaign 
$^{2}$Sandia National Laboratories \\
\texttt{\{xiaonaz2, lourent2\}@illinois.edu} \\[5pt]
\href{https://plan-lab.github.io/VisAnom}{\texttt{\textcolor{IllinoisOrange}{plan-lab.github.io/VisAnom}}}}

\begin{document}

\maketitle

\begin{abstract}
Recent advances in Vision--Language Models (VLMs) have achieved impressive performance across many tasks, yet prior studies report unsatisfactory performance when applying large language or multimodal models to finding abnormal patterns in sequential data. 
Public anomaly detection benchmarks typically provide interval annotations but not natural-language rationales, making it difficult to fine-tune VLMs to produce grounded, interpretable decisions.
To address this gap, we construct \textbf{\dataset{}}, a curated benchmark built from public time-series datasets and augmented with high-quality anomaly explanations selected from multiple large VLMs using fine-grained, task-specific rewards.
Through fine-tuning on this benchmark, we develop \modelname{}, a parameter-efficient VLM for time-series anomaly detection.
Results on~\datasetnc{} show that \modelnamenc{} achieves more accurate anomaly localization and consistently outperforms all baselines, with improvements of at least 21.23 and 23.87 percentage points in precision and F\textsubscript{1}, respectively. Additional experiments on the \textsc{TSB-AD-U} benchmark demonstrate strong cross-benchmark generalization, with \modelnamenc{} improving precision and F\textsubscript{1} by 9.57 and 13.39 percentage points, respectively.
\end{abstract}

\input{sections/01_introduction}

\begin{figure}[t]
\centering
\includegraphics[width=0.99\textwidth]{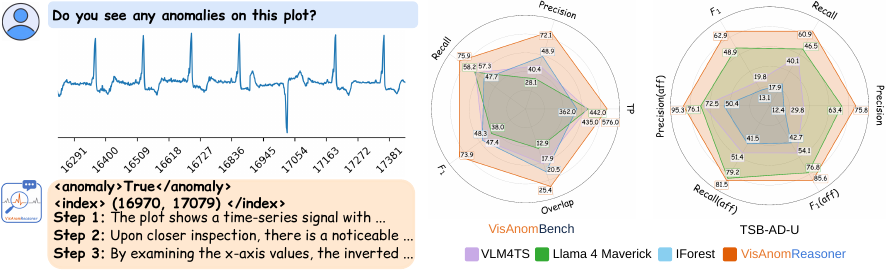}
\vspace{-0.2cm}
\caption{Given a time series plot (left top), \modelname{} locates anomalies while providing details grounded in the plot (left bottom).
Experimental results on two benchmarks demonstrate \modelnamenc{} outperforms the strongest baselines by large margins across all metrics (right).}
\vspace{-0.3cm}
\label{fig:teaser}
\end{figure}

\input{sections/02_related_work}

\input{sections/03_benchmark}
\input{sections/03_method}

\input{sections/04_experiments}

\input{sections/05_conclusion}

\begin{ack}
This work was supported by the U.S. Defense Advanced Research Projects Agency (DARPA) under award HR001125C0303, the U.S. Department of Energy, National Nuclear Security Administration, Office of Defense Nuclear Nonproliferation Research and Development, and by the Laboratory Directed Research and Development program at Sandia National Laboratories.
Sandia National Laboratories is a multi-mission laboratory managed and operated by National Technology \& Engineering Solutions of Sandia, LLC (NTESS), a wholly owned subsidiary of Honeywell International Inc., for the U.S. Department of Energy’s National Nuclear Security Administration (DOE/NNSA) under contract DE-NA0003525. This written work is authored by an employee of NTESS. The employee, not NTESS, owns the right, title and interest in and to the written work and is responsible for its contents. Any subjective views or opinions that might be expressed in the written work do not necessarily represent the views of the U.S. Government. The publisher acknowledges that the U.S. Government retains a non-exclusive, paid-up, irrevocable, world-wide license to publish or reproduce the published form of this written work or allow others to do so, for U.S. Government purposes. The DOE will provide public access to results of federally sponsored research in accordance with the DOE Public Access Plan.
\end{ack}

\bibliographystyle{abbrv}
\bibliography{neurips}

\newpage
\appendix
\input{sections/06_appendix}
\newpage
\input{checklist.tex}
\end{document}

%% file: sections/01_introduction.tex
\section{Introduction}
\label{sec:introduction}

Detecting abnormal patterns in time-series data is a critical problem in applications such as industrial process monitoring, healthcare diagnostics, and cyber--physical systems, where timely and interpretable decisions are essential~\citep{blazquez2021review, garg2021evaluation, sgueglia2022systematic, yang2023deep, soleimani2023effectiveness}. Beyond identifying when anomalies occur, practitioners often require explanations that justify why a region is anomalous, \eg to diagnose failure modes and support downstream decision-making. However, existing time-series anomaly detection methods focus exclusively on numeric scores or binary labels~\citep{paparrizos2022tsb, zamanzadeh2024deep, schmidl2022anomaly, zhou2026mTSBench}, offering limited interpretability and little insight into why a particular interval was flagged as anomalous.\looseness-1

Recent advances in Vision--Language Models (VLMs) have demonstrated strong multimodal reasoning capabilities across tasks such as image captioning~\citep{li2023blip, wang2022ofa}, visual question answering~\citep{liu2023visual, kim2024image, ogunleye2025ecvlm, susladkar2026pyratok, shen2026cogniroute}, chart understanding~\citep{huang2025evochart, yang2025effective, pmlr-v306-susladkar26a}, and visual instruction following~\citep{li2025robotic, pchelintsev2025lera, liu2026palm}. These models can align visual evidence with natural language explanations~\citep{rasheed2024glamm, zhang2024groundhog, wahed2027prima, Li_2026_CVPR} and generate step-by-step reasoning in complex visual domains~\citep{zhang2024multimodalcot, shao2024visualcot, li2026supersensing, shen2025finegrained, tabassum-etal-2025-mmplanner, yu2025core3d}. Despite this progress, applying VLMs to time-series anomaly detection remains challenging. Time-series plots lack explicit object boundaries and spatial structure, requiring models to reason over temporal patterns, periodicity, and deviations rather than discrete visual entities. As a result, general-purpose VLMs applied out of the box often produce coarse, unstable anomaly predictions and ungrounded explanations when faced with time-series data~\citep{xu2025can, zhoucan, wang2025can}.\looseness-1

A key obstacle to adapting VLMs for time-series anomaly reasoning is the lack of suitable supervision.
Public time-series datasets typically provide only anomaly intervals or point-wise labels~\citep{paparrizos2022tsb, zamanzadeh2024deep, schmidl2022anomaly, zhou2026mTSBench}, without accompanying explanations, which significantly limits the application of supervised fine-tuning.
As a result, existing approaches either adopt task-specific detection pipelines without explicit reasoning supervision~\citep{he2025harnessing}, or attempt to elicit anomaly reasoning from pretrained models through prompting alone~\citep{parkdelving, liu2025large, zhoucan}, leading to limited anomaly localization accuracy in practice. 
Most existing approaches operate on textual or short time-series inputs and are limited by context window constraints, while only one method supports long time-series inputs.

This work addresses this gap by introducing \textbf{\dataset{}}, a benchmark for vision--language time-series anomaly reasoning.
\datasetnc{} is constructed from multiple public benchmarks spanning diverse domains and anomaly types, and augments them with temporally grounded natural-language explanations aligned with anomaly intervals.
It enables supervised learning for both interval-level anomaly localization and explanation generation directly from time-series plots.

We introduce \modelname{}, a parameter-efficient VLM fine-tuned on \datasetnc{} for reasoning-based time-series anomaly detection. Experimental results show that explanation-augmented supervision enables \modelnamenc{} to achieve more accurate anomaly localization and improved temporal alignment compared to both large general-purpose VLMs and specialized large models across multiple evaluation metrics. The contributions are summarized as follows:

\begin{itemize}[itemsep=0.5ex, parsep=0pt, topsep=0pt, leftmargin=0.7cm]

     \item[\textbf{(1)}] We formulate time-series anomaly detection as a plot-grounded vision-language reasoning task that jointly requires interval-level localization and structured explanation generation. This formulation moves beyond scalar anomaly scores by evaluating whether models can align visual evidence, temporal boundaries, and natural-language reasoning in a single prediction.

    \item[\textbf{(2)}] We curate \dataset{}, the first explanation-augmented benchmark for vision-language time-series anomaly reasoning spanning diverse domains and anomaly types. Unlike existing benchmarks that provide only point-wise or interval labels, \datasetnc{} pairs anomaly intervals with temporally grounded natural-language rationales, enabling models to learn not only where anomalies occur but also why they are visually and temporally abnormal.

     \item[\textbf{(3)}] We introduce \modelname{}, a compact vision-language anomaly reasoning model, and evaluate it against 15 baselines. Despite its size, \modelnamenc{} consistently outperforms general-purpose VLMs, specialized LLM/VLM anomaly detectors, time-series foundation models, and classical detectors, improving precision and F\textsubscript{1} by at least 21.23 and 23.87 pp on \datasetnc{} and achieving strong cross-benchmark generalization on \textsc{TSB-AD-U}.

     \item[\textbf{(4)}] We validate the supervision strategy through ablations, human evaluation, and robustness analysis. Reasoning supervision improves F\textsubscript{1} over interval-only fine-tuning from 61.87\% to 73.94\%, while reward-selected multi-model supervision adds 9.13 points over single-model supervision. In a blinded human study, \modelnamenc{} outperforms the Base model in 92.9\% of paired evaluations. Controlled experiments also show robustness to plot-rendering variations.
\end{itemize}

%% file: sections/02_related_work.tex
\section{Related Work}
\label{sec:related_works}

\textbf{LLMs for Time-Series Reasoning and Modeling.}
Recent work has explored using large language models for time-series reasoning and question answering by encoding time series as text or symbolic representations. ChatTS~\citep{xie2024chatts} and AXIS~\citep{lan2025axis} study conversational analysis and Q\&A over time-series data, while Time-MQA~\citep{kong2025time} introduces a multi-task benchmark that includes anomaly-related questions. Time-RA~\citep{yang2025time} uses short time series given as text and plots to predict fine-grained anomaly categories with explanations, but does not evaluate anomaly localization. A parallel line of work adapts LLMs for time-series forecasting and classification by modeling sequences as text or tokens~\citep{luo2025time, wang2025can, zhang2025tempogpt, zhang2025timemaster, wang2025chattime, gruver2023llmtime}. These approaches are constrained by context window limits and cannot scale to long time series with thousands of observations without preprocessing, which introduces additional challenges such as information loss and fragmented temporal context. Consequently, existing LLM-based time-series methods offer limited support for real-world anomaly detection settings that demand long-horizon analysis, accurate localization of anomaly intervals, and interpretable, visually grounded explanations.

\textbf{VLMs and Multimodal Models for Anomaly Detection.}
An emerging direction addresses the limitations of text-based time-series input by operating directly on visual representations, where plots expose trends, periodicity, and deviations that naturally support anomaly detection via visual reasoning. Early work in this direction encodes short time-series windows with a vision encoder and relies on large VLMs, such as GPT-4o, to identify anomalous regions~\citep{he2025harnessing}.
Complementary studies benchmark open-source and proprietary multimodal models on time-series anomaly detection tasks, showing that out-of-the-box performance remains inconsistent and often unstable~\citep{xu2025can}.
Other approaches explore prompt-based strategies to guide anomaly detection, for example by injecting explicit time-index information~\citep{liu2025large, parkdelving, zhoucan}, but these methods are typically constrained to short sequences due to context window limitations. These works highlight the promise of visual reasoning for time-series anomaly detection, but existing benchmarks do not support models that jointly perform interval-level localization and produce plot-consistent explanations. To this end, we introduce \datasetnc{}, an explanation-augmented benchmark that enables learning through reasoning traces, allowing models to jointly learn accurate anomaly localization and plot-consistent explanations.

%% file: sections/03_benchmark.tex
\section{\datasetnc{} Time-Series Anomaly Reasoning Benchmark}
Existing benchmarks for time-series anomaly detection typically provide only anomaly locations, without explanations describing how or why the labeled regions are anomalous~\citep{paparrizos2022tsb, zhou2026mTSBench, WenigEtAl2022TimeEval, wu2021current}.
This limits their suitability for task-specific adaptation of vision--language models, which require explanatory supervision for better contextual explanation of anomalies.
To address this limitation, we introduce \dataset{}, which augments four public time-series anomaly detection benchmarks with temporally grounded natural-language explanations aligned with annotated anomaly intervals.

For each input pair \((I(x), C)\), where \(I(x)\) is a time-series plot and \(C\) denotes optional contextual information, a VLM produces a structured output consisting of:
(1)~a set of predicted anomaly intervals, $A$,  enclosed by an \texttt{<index>} tag,
and
(2)~a step-by-step explanation,
$E\!=\!\{\epsilon_k\}_{k=1}^{K},
$
where each \(\epsilon_k\) is a numbered reasoning statement (\eg \texttt{Step 1: ...}), enclosed by a \texttt{<think>} tag.

What constitutes an anomaly varies across domains and application contexts. For instance, brief spikes may be expected in noisy telemetry data, while gradual trend deviations or prolonged level shifts are of primary interest.
To account for this variability, ground-truth anomaly intervals are provided during explanation elicitation, guiding models to produce explanations that are temporally grounded in the visualization and aligned with the intended semantic definition of abnormality. Further details on dataset construction are provided in Appendix \ref{apd:benchmark}.

\datasetnc{} is constructed in four stages: (i) segment public time series into plot-renderable windows with valid anomaly intervals; (ii) render each segment as an image with axis labels and optional context; (iii) elicit structured anomaly decisions and reasoning traces from multiple large VLMs; and (iv) select the highest-quality candidate using a reward that combines interval accuracy, visual groundedness, axis awareness, and clarity.

\subsection{Reward-Guided Explanation Selection}

For each time series, we construct a pool of candidate structured outputs using four general-purpose VLMs. Each generated output \((A, E)\) is evaluated using a composite reward function
\begin{equation}
\setlength{\abovedisplayskip}{6pt}
\setlength{\belowdisplayskip}{6pt}
\begin{aligned}
\mathcal{R}(A, E) =\;&
\lambda_{\text{ano}} \mathcal{S}_{\text{ano}}(A, \mathcal{A}^\star)
+ \lambda_{\text{vis}} \mathcal{S}_{\text{vis}}(E)  + \lambda_{\text{axi}} \mathcal{S}_{\text{axi}}(E)
+ \lambda_{\text{cla}} \mathcal{S}_{\text{cla}}(E),
\end{aligned}
\label{eq:reward}
\end{equation}
where $\{\lambda_{\text{ano}}, \lambda_{\text{vis}}, \lambda_{\text{axi}}, \lambda_{\text{cla}} \}$ are weighting coefficients.

\noindent\faChartLine~\textbf{Anomaly Accuracy.}
$\mathcal{S}_{\text{ano}}(A, \mathcal{A}^\star)$ is a length-weighted, range-based F\textsubscript{1} score that measures temporal overlap between predicted and ground-truth anomaly intervals. It rewards coverage of true anomaly ranges while penalizing missed anomalies and overextended predictions.

\noindent\faChartLine~\textbf{Visual Groundedness.}
$\mathcal{S}_{\text{vis}}(E)$ measures whether the reasoning explicitly references temporal patterns directly observable in the plot, such as spikes, drops, or level shifts.

\noindent\faChartLine~\textbf{Axis Awareness.}
$\mathcal{S}_{\text{axi}}(E)$ evaluates whether timestamps, indices, and value ranges in the reasoning are consistent with the plot axes, penalizing hallucinated or unsupported numerical claims.

\noindent\faChartLine~\textbf{Clarity.}
$\mathcal{S}_{\text{cla}}(E)$ assesses whether the reasoning is logically ordered, non-redundant, and clearly links visual observations to the final anomaly decision in a coherent manner.

For each time series, we retain only the candidate with the highest reward as the supervision target for supervised fine-tuning of \modelnamenc{}. Additional details are provided in Appendix \ref{apd:rating}.

\subsection{Dataset Composition and Statistics}
\input{assets/tables/VisAnom-Bench}
Time series in \datasetnc{} are drawn from four public anomaly detection benchmarks: \textsc{KPI}~\citep{xu2018unsupervised}, \textsc{GutenTAG}~\citep{WenigEtAl2022TimeEval}, \textsc{UCR-EGI}~\citep{gao2020ensemble}, and \textsc{UCR-TSAD}~\citep{wu2021current}. 
The collection comprises both real-world and synthetic time series, covering diverse domains, anomaly types, and temporal characteristics.
Overall dataset statistics are summarized in Table~\ref{tab:dataset_stats}.
In total, \datasetnc{} contains 2{,}576 training time series and 740 held-out test time series, with a strictly disjoint test set used exclusively for evaluation. 

%% file: assets/tables/VisAnom-Bench.tex
\begin{wraptable}{r}{0.625\textwidth}
\vspace{-0.4cm}
\centering
\scriptsize
\caption{
\textbf{Time-series statistics in \datasetnc{}.}
\#TS: number of time series; \#Dom: number of domains; Avg. Len: average length; AR(\%): anomaly ratio; Expl. Len.: average number of words in an explanation.
}
\label{tab:dataset_stats}
\vspace{-0.2cm}
\resizebox{\linewidth}{!}{
\begin{NiceTabular}{l c c c c  c}[colortbl-like]
    \toprule
    \rowcolor[HTML]{EFEFEF}
    \textbf{Benchmark} &
    \textbf{\#TS} &
    \textbf{\#Dom} &
    \textbf{Avg. Len} &
    \textbf{AR(\%)} &
    \textbf{Expl. Len.}\\
    \midrule

    \textsc{KPI}~\citep{xu2018unsupervised}
    & 160 & 1 & 1,777 & 4.3 & 170\\

    \rowcolor[HTML]{F9F9F9}
    \textsc{GutenTAG}~\citep{WenigEtAl2022TimeEval}
    & 810 & 10 & 5,000 & 2.2 & 114\\

    \textsc{UCR-EGI}~\citep{gao2020ensemble}
    & 2,097 & 5 & 8,268 & 5.8 & 112\\

    \rowcolor[HTML]{F9F9F9}
    \textsc{UCR-TSAD}~\citep{wu2021current}
    & 249 & 8 & 2,755 & 6.9 & 112\\

    \bottomrule
\end{NiceTabular}
}
\vspace{-0.3cm}
\end{wraptable}

%% file: sections/03_method.tex
\section{\modelnamenc{} Model}
\label{sec:sft}

\modelnamenc{} is a parameter-efficient VLM designed for reasoning-based time-series anomaly detection from plots. Unlike prior approaches that rely on prompt engineering to detect anomalies, \modelnamenc{} directly predicts anomaly intervals and generates grounded explanations. 
Prior approaches~\citep{zhoucan, parkdelving, liu2025large} often depend on extensive prompt engineering and external proprietary models to infer anomalies from visual inputs. In contrast, \modelnamenc{} leverages explanation-augmented supervision from \datasetnc{}, where each training example specifies both target anomaly intervals and a preferred reasoning trace aligned with visual evidence.\looseness-1

Consider a univariate time series \(x=\{x_1,\ldots,x_T\}\in\mathbb{R}^T\), where \(x_t\) denotes the observed value at timestamp \(t\).
Let the ground-truth anomalies be a set of \(m\) temporal intervals
$\mathcal{A}^\star=\{(s_i^\star,e_i^\star)\}_{i=1}^{m}$, $1\le s_i^\star<e_i^\star\le T$.
The objective of time-series anomaly detection is to localize anomalous intervals by predicting a set \(A=\{(s_i,e_i)\}_{i=1}^{\hat m}\). We formulate time-series anomaly detection as a vision--language reasoning task that jointly localizes anomalies and explains them from a plotted rendering of the series.
Given a time-series plot \(I(x)\) and optional contextual information \(C\), a vision--language model \(\mathcal{M}_\theta\) outputs predicted anomaly intervals \(A\) and a natural-language explanation \(E\):
\begin{equation}
\setlength{\abovedisplayskip}{6pt}
\setlength{\belowdisplayskip}{6pt}
(I(x),C)\xrightarrow{\;\mathcal{M}_\theta\;}(A,E),
\end{equation}
where \(E\) is grounded in visual evidence, such as spikes, level shifts, or periodic deviations.

Unlike conventional anomaly detection methods that output anomaly scores or point-wise labels~\citep{schmidl2022anomaly}, and prior LLM- and VLM-based approaches that cast anomaly detection as classification or question answering~\citep{xie2024chatts, lan2025axis, zhang2025tempogpt}, our formulation explicitly couples interval-level localization with an interpretable explanation \(E\) grounded in observable visual patterns. We optimize \(\mathcal{M}_\theta\) using supervised fine-tuning of Qwen2.5-VL-3B and Qwen2.5-VL-7B base models~\citep{bai2025qwen2} on our curated \datasetnc{} training set, resulting in the 3B and 7B variants of \modelnamenc{}, respectively.
During training, the model is supervised to generate structured outputs that include \texttt{<anomaly>} tags for anomaly decisions, \texttt{<index>} tags for interval localization, and \texttt{<think>} tags for generated reasoning traces, jointly supervising detection, localization, and reasoning.

%% file: sections/04_experiments.tex
\section{Experiments}\label{sec:experiments}

We evaluate \modelnamenc{}'s ability to localize anomalies in time-series plots and generate interpretable explanations, in comparison to baseline models. 

\textbf{Datasets.}
We conduct experiments on (1) the test portion of \datasetnc{} and (2) \textsc{TSB-AD-U}~\citep{liu2024elephant}, a widely used benchmark for anomaly detection with time series from various domains.

\textbf{Baselines.}
We benchmark \modelnamenc{} against five categories of baseline models. For both general large VLMs and small VLMs, we use the same prompting setup as in \modelnamenc{} to ensure a fair comparison. Detailed descriptions of all baseline models are provided in Appendix \ref{apd:models}.

\noindent \tightcolorbox{genVLM}{\textbf{General Large VLMs.}} This category includes powerful frontier general-purpose VLMs, such as Grok-4-Fast~\citep{grok2024} and LLaMA~4~Maverick~\cite{meta-llama4-2025}.

\noindent \tightcolorbox{smVLM}{\textbf{Small VLMs.}} These are lightweight open-source models such as Qwen2.5-VL-7B~\citep{bai2025qwen2}, Idefics3-8B~\citep{laurencconbuilding}, SmolVLM-2.2B~\citep{marafioti2025smolvlm}, and LLaVA-7B~\citep{liu2023visual}.

\noindent \tightcolorbox{FM}{\textbf{Foundation Models.}} This group contains foundation models proposed for time-series anomaly detection or time-series analysis more broadly, such as TimesFM~\citep{das2024decoder} and Chronos~\citep{ansarichronos}.

\noindent \tightcolorbox{specLM}{\textbf{Specialized Large Models.}} This baseline set comprises models developed or adapted for time-series anomaly detection, which are AnomLLM~\citep{zhoucan}, LLM-TSAD~\citep{parkdelving}, LLMAD~\citep{liu2025large}, and VLM4TS~\citep{he2025harnessing}.

\noindent \tightcolorbox{detector}{\textbf{Classical Detectors.}} This category comprises established anomaly detection methods from the time-series literature, such as Sub-PCA~\citep{aggarwal2016outlier}, Matrix Profile~\citep{linardi2020matrix}, and Isolation Forest (IForest)~\citep{liu2008isolation}.

\textbf{Evaluation Metrics.}
For \datasetnc{}, anomaly detection performance is evaluated at the interval level using Precision, Recall, and F\textsubscript{1} score metrics \cite{baireddy2021spacecraft}.
A predicted interval is counted as a true positive (TP) if it overlaps with a ground-truth interval (if multiple predicted intervals overlap with the same ground-truth interval, only one TP is counted); a predicted interval is counted as a false positive (FP) if it does not overlap with any ground-truth interval; a ground-truth interval is counted as a false negative (FN) if it does not overlap with any predicted interval. 
The Overlap score quantifies the temporal alignment between predicted and ground-truth anomaly intervals.
It penalizes both under-coverage and over-extension of predicted intervals relative to the ground truth.\looseness-1

For the \textsc{TSB-AD-U} benchmark, following established practice~\citep{zhoucan}, we report both \emph{standard} metrics (including Precision, Recall, and F\textsubscript{1}) and respective \emph{affiliation} metrics~\citep{huet2022local}. 
Standard metrics emphasize point-wise correctness and impose strict penalties for boundary misalignment.
In contrast, affiliation metrics operate at the event level and emphasize temporal association between predicted and ground-truth anomaly intervals, allowing tolerance to boundary imprecision.
Additional details about the evaluation metrics are provided in Appendix~\ref{apd:imple}.

\subsection{Experimental Results}
\label{sec:results}

\subsubsection{Results on \datasetnc{}}

Table~\ref{tab:anom_results} reports interval-level anomaly detection performance on \datasetnc{}.
Across all evaluation metrics, \modelnamenc{} consistently outperforms general large VLMs, small VLMs, specialized large models, foundation models, and classical anomaly detectors.
Both the 3B and 7B variants outperform all baselines in Precision, Recall, F\textsubscript{1}, and Overlap scores, indicating strong performance in both anomaly identification and temporal localization.

\textbf{Precision--Recall Trade-offs.}
Many baseline methods achieve higher recall by flagging large portions of the time series as anomalous, leading to substantially inflated FP counts.
In particular, large general-purpose VLMs such as Grok-4-Fast~\citep{grok2024} and LLaMA-4-Maverick~\citep{meta-llama4-2025} recover a sizable fraction of true anomalies, with recall values of 31.88\% and 58.23\%.
However, this recall is achieved at the cost of excessive FPs: these models produce at least 837 FPs, which is more than twice the number of true anomalies they correctly identify.
As a result, precision remains below 28.15\%, and F\textsubscript{1} scores do not exceed 37.96\%.
This over-flagging behavior is further reflected in low overlap scores of 3.26\% and 12.92\%, which are at least 14.15 percentage points (pp) lower than those achieved by the \modelnamenc{} 3B variant, indicating that predicted anomaly intervals are often overly long or weakly aligned with true anomaly boundaries.

Small VLMs, including Qwen2.5-VL-7B~\citep{bai2025qwen2}, Idefics3-8B~\citep{laurencconbuilding}, and SmolVLM-2.2B~\citep{marafioti2025smolvlm}, further exacerbate this issue.
They all suffer from low precision and overlap due to aggressive over-flagging and limited reasoning capacity.
For example, Qwen2.5-VL-7B, the strongest baseline in this category, correctly identifies 517 anomaly intervals but also produces 1,784 FPs.
This over-detection substantially degrades precision and boundary quality, yielding precision, F\textsubscript{1}, and overlap scores that are 49.62~pp, 40.15~pp, and 11.31~pp lower than those for \modelnamenc{} (7B), respectively.
\input{assets/tables/tab_main_results}

Foundation models such as TimesFM~\citep{das2024decoder} and Chronos~\citep{ansarichronos}, which adapt forecasting error for anomaly detection, exhibit fewer FPs (approximately 500).
However, this behavior is largely driven by the evaluation protocol, which flags only the top 5\% of points with the highest anomaly scores.
Even under this favorable constraint, their precision remains roughly 20~pp lower than that of classical anomaly detectors such as Sub-PCA~\citep{aggarwal2016outlier}, Matrix Profile~\citep{linardi2020matrix}, and IForest~\citep{liu2008isolation}, which are evaluated under the same protocol.
Among classical methods, IForest stands out as the strongest traditional baseline, achieving the second-best result on four of the seven reported metrics.

Among specialized large models, most approaches rely on different prompting strategies with GPT-4o and consistently exhibit oversensitivity to temporal fluctuations~\citep{zhoucan, parkdelving, liu2025large}.
LLM-TSAD is particularly affected, producing over 5{,}000 FPs, indicating severe over-flagging and resulting in precision as low as 7.82\%. 
VLM4TS, the strongest specialized model for time-series anomaly detection, detects 129 fewer true anomaly intervals and misclassifies 402 more normal intervals as anomalous compared to the \modelnamenc{} 3B variant, resulting in over 20~pp lower precision and F\textsubscript{1} score. In summary, \modelnamenc{} variants detect the largest number of anomaly intervals while producing the fewest FP predictions, with the 7B variant achieving at least 23.17~pp and 25.64~pp higher precision and F\textsubscript{1} score, respectively, than any of the baseline methods.

\noindent \textbf{Temporal Boundary Localization.}
The Overlap score quantifies temporal boundary localization by penalizing both missed coverage and excessive extension of predicted anomaly intervals.
As shown in the last column of~\cref{tab:anom_results}, this criterion is challenging: 10 methods score below 15\%, and seven fall below 5\%, indicating widespread difficulty in localizing anomaly boundaries.
In contrast, the \modelnamenc{} variants achieve the highest overlap scores, exceeding all baselines by at least 6.54~pp and 4.82~pp for the 3B and 7B variants, respectively.
In particular, \modelnamenc{} (7B) attains an overlap improvement of 11~pp compared to Qwen2.5-VL-7B~\citep{bai2025qwen2}, and at least 12.43~pp over the large general-purpose VLMs~\citep{grok2024, meta-llama4-2025}, demonstrating substantially tighter temporal alignment.

\input{assets/tables/TSB-U-AD}

Comparing LLM-TSAD~\citep{parkdelving} and IForest~\citep{liu2008isolation} shows that overlap measures boundary quality rather than anomaly volume.
IForest detects only 362 true anomaly intervals yet achieves an overlap score 0.90~pp higher than LLM-TSAD, which identifies 477 true anomalies along with 5621 FPs.
Since the evaluation set contains 740 time series, LLM-TSAD predicts more than seven anomaly intervals per series on average.
These excessive predictions are heavily penalized by the overlap metric, yielding weaker boundary alignment than a far more conservative detector.

\subsubsection{Results on \textsc{TSB-AD-U}}

Results in \cref{tab:tsbadu} compare \modelnamenc{} with existing methods across six evaluation metrics on the public benchmark \textsc{TSB-AD-U}, which contains time series with unseen characteristics and anomaly definitions. Compared to large VLMs, such as LLaMA-4-Maverick~\citep{meta-llama4-2025}, \modelnamenc{} (7B) performs best
on all six metrics, with precision and F\textsubscript{1} improving by about 10~pp or more. This result indicates that fine-tuning on \datasetnc{} enables \modelnamenc{} to learn transferable anomaly detection capabilities that generalize to novel time-series patterns, reaching performance better than or comparable to a large model despite its substantially smaller scale.

Qwen2.5-VL-7B~\citep{bai2025qwen2} attains the highest standard precision (66.18\%) and standard F\textsubscript{1} (49.52\%) among all baselines, as well as the highest standard recall (48.85\%) among the small VLM baselines. Nevertheless, all of these values are improved by \modelnamenc{} (7B) (by 9.57~pp, 12.06~pp, and 13.39~pp for standard precision, recall, and  F\textsubscript{1}, respectively). The advantage of \modelnamenc{} (7B) over Qwen2.5-VL-7B in affiliation precision, recall, and F\textsubscript{1} is also substantial (by 19.37~pp, 6.58~pp, and 11.22~pp, respectively). The largest improvement is in affiliation precision, which indicates that \modelnamenc{} avoids excessive flagging of anomalies and thereby reduces FPs.

A similar pattern is observed for the foundation models TimesFM~\citep{das2024decoder} and Chronos~\citep{ansarichronos}. While both models obtain high standard and affiliation recall, the best of them (Chronos) has standard and affiliation precision approximately 30~pp and 20~pp lower than those of \modelnamenc{}, respectively. This large precision gap indicates systematic over-flagging, which can overwhelm users with false alarms and limit practical usefulness despite strong recall performance.

Performance of specialized large models that rely on GPT-4o for decision making varies widely, with AnomLLM typically being the worst one among these baselines. Still, all of them underperform relative to \modelnamenc{} across all six metrics. Notably, they mostly lag behind large general-purpose VLMs such as LLaMA-4-Maverick~\citep{meta-llama4-2025}, which are evaluated under the same prompting strategy as \modelnamenc{}. This observation raises questions about the effectiveness of task-specific prompting strategies proposed by these methods~\citep{zhoucan, parkdelving, liu2025large}. 

Classical anomaly detectors perform worst overall, highlighting their limited robustness to the diverse time-series characteristics present in \textsc{TSB-AD-U}, a limitation widely observed in prior work~\citep{schmidl2022anomaly}.

\modelnamenc{} delivers strong and well-balanced performance across both standard and affiliation metrics, providing empirical evidence that explanation-augmented supervision yields improved generalization and more reliable event-level anomaly detection compared to existing approaches.

\input{assets/figures/qual_visanom}

\section{Qualitative Analysis}

\cref{fig:qual} presents a qualitative comparison of anomaly detection and reasoning across models on a representative time-series example.
VLM4TS~\citep{he2025harnessing}, a ViT+GPT-4o model designed for time-series anomaly detection, identifies two short intervals and one point as anomalous but provides limited justification and fails to align with the true anomaly region.
The predicted intervals correspond to normal variations within the underlying periodic structure, indicating sensitivity to local fluctuations rather than meaningful pattern deviations in the overall temporal behavior.
LLMAD~\citep{liu2025large}, a prompting-based method that supports anomaly explanations, predicts as many as 18 anomaly intervals, but only one overlaps with the ground-truth interval. Moreover, the predicted interval covers only about 3\% of the true anomalous region, and the accompanying explanation is not aligned with the actual temporal trend in the time-series plot. LLaMA~4~Maverick~\cite{meta-llama4-2025} exhibits similar behavior, predicting eight anomalous points, none of which intersect the ground-truth anomaly region; its explanation remains largely generic and does not reference the observed trend in the time series.\looseness-1 

In contrast, \modelnamenc{} produces a tighter anomaly interval that closely matches the ground truth and generates a coherent, step-by-step reasoning trace that links changes in amplitude and pattern to observable deviations in the signal.
This illustrates that explanation-augmented supervision enables \modelnamenc{} to achieve more precise localization and better-grounded reasoning than both customized and larger general-purpose VLMs. Additional analysis is in Appendix~\ref{apd:qual_ex}.

\section{Ablation Studies}\label{sec:ablation}

\input{assets/figures/base_vs_SFT}
\paragraph{Effect of Supervised Fine-Tuning.}
\cref{fig:base_vs_SFT} compares the \modelnamenc{} 3B and 7B variants against their respective Qwen2.5-VL base models~\citep{bai2025qwen2} on \datasetnc{}, with percentage gains annotated above each bar.
Supervised fine-tuning improves all metrics, with the largest gains in precision, amounting to 180\% for 3B and 220\% for 7B, which reflects a significant reduction in FPs.
In contrast, recall improves only marginally, indicating that fine-tuning primarily reduces FPs rather than substantially increasing anomaly coverage.
This pattern is consistent with explanation-augmented supervision emphasizing context-aware discrimination, helping the model distinguish true anomalies from normal variations.
Improvements in overlap indicate more accurate temporal boundary localization.
These results demonstrate the effectiveness of explanation-augmented supervision, particularly for smaller models, in improving both detection quality and localization accuracy.

\begin{wraptable}{r}{0.6\textwidth}
\vspace{-0.35cm}
\centering
\scriptsize
\setlength{\tabcolsep}{4pt}
\caption{
\textbf{Fine-tuning with explanation improves precision and recall over interval-only supervision.}}
\label{tab:reasoning_ablation}
\vspace{-0.15cm}
\resizebox{\linewidth}{!}{
\begin{tabular}{lcccccc}
\toprule
\textbf{Mode} & \textbf{TP} & \textbf{FP} & \textbf{FN} & \textbf{Prec.} & \textbf{Rec.} & \textbf{F\textsubscript{1}} \\
\midrule
Base & 517 & 1784 & 242 & 22.47 & 68.12 & 33.79 \\
No reasoning & 516 & 393 & 243 & 56.77 & 67.98 & 61.87 \\
With reasoning & 576 & 223 & 183 & 72.09 & 75.89 & 73.94 \\
\bottomrule
\end{tabular}
}
\vspace{-0.3cm}
\end{wraptable}
\textbf{Effect of Reasoning Supervision.}
We evaluate whether the gains come specifically from reasoning supervision rather than from task-specific fine-tuning alone by comparing three settings: (i) the off-the-shelf pretrained base model, (ii) fine-tuning to predict anomaly intervals only, and (iii) fine-tuning to predict both anomaly intervals and reasoning traces.
As shown in~\cref{tab:reasoning_ablation}, interval-only fine-tuning already reduces FPs substantially, increasing precision from 22.47\% to 56.77\%, but does not recover additional missed anomalies.
In contrast, adding reasoning traces further improves both precision and recall to 72.09\% and 75.89\%, respectively, raising F1 from 61.87\% to 73.94\%. These results show that explanation-augmented supervision provides gains beyond task-specific interval fine-tuning alone.

\noindent\textbf{Effect of multi-model supervision.}
Because \datasetnc{} is constructed from VLM-generated candidate explanations, we examine whether supervision selected from multiple generators is more effective than relying on a single model. We compare training on \datasetnc{}, where candidate explanations are generated by four general-purpose VLMs and selected by a separate judge using our task-specific reward, against training on explanations generated only by Gemma-3-27B-IT~\citep{gemma_2025}, under the same downstream training setup. Single-model supervision yields 535 TP, 357 FP, and 224 FN, corresponding to 59.98\% precision, 70.49\% recall, and 64.81\% F\textsubscript{1}. In contrast, training on \datasetnc{} achieves 576 TP, 223 FP, and 183 FN, improving precision, recall, and F\textsubscript{1} to 72.09\%, 75.89\%, and 73.94\%, respectively. This corresponds to gains of 12.11, 5.40, and 9.13 percentage points, together with fewer false positives and false negatives. These results suggest that reward-selected supervision from multiple generators provides more effective training targets than supervision from a single generator, especially by reducing false-positive errors substantially.

\begin{wrapfigure}{r}{0.53\textwidth}
\vspace{-0.4cm}
\centering
\begin{tikzpicture}
\begin{axis}[
    xbar stacked,
    width=0.53\textwidth,
    height=2.35cm,
    xmin=0, xmax=100,
    symbolic y coords={Preference},
    ytick=data,
    yticklabels={},
    xtick={0,25,50,75,100},
    xticklabel style={font=\scriptsize},
    bar width=16pt,
    axis line style={draw=none},
    tick style={draw=none},
    xmajorgrids,
    grid style={gray!20},
    legend style={
        at={(0.5,1.25)},
        anchor=south,
        legend columns=3,
        font=\scriptsize,
        draw=none
    },
    nodes near coords={
\pgfmathprintnumber[fixed,precision=1]{\pgfplotspointmeta}\%
},
    nodes near coords style={font=\scriptsize\bfseries, text=black},
]
\addplot+[fill=orange!60, draw=white] coordinates {(69.6,Preference)};
\addplot+[fill=blue!45, draw=white] coordinates {(29.7,Preference)};
\addplot+[fill=gray!45, draw=white] coordinates {(0.7,Preference)};
\legend{\modelname{} Win,  Base Win, Tie}
\end{axis}
\end{tikzpicture}
\vspace{-0.15cm}
\caption{\textbf{Explanation win-rate} between base model and \modelnamenc{}.}
\label{tab:explanation_preference}
\vspace{-0.35cm}
\end{wrapfigure}

\noindent\textbf{Quality of Explanation.} We evaluate the explanation quality by conducting a paired comparison on the 740 test time series in~\datasetnc{}.
For each time series, GPT-4o is given the plot and two explanations, one generated by the base model (Qwen2.5-VL-7B) and one by \modelnamenc{} (7B), and selects the better explanation based on visual groundedness, axis consistency, and clarity.
As shown in~\cref{tab:explanation_preference}, \modelnamenc{} is preferred in 69.6\% of cases, compared with 29.7\% for the base model and 0.7\% ties.

\begin{wraptable}{r}{0.45\textwidth}
\vspace{0.1cm}
\centering
\scriptsize
\setlength{\tabcolsep}{4pt}
\caption{
\textbf{Human evaluation of explanation quality on a 1--5 scale.}}
\label{tab:human_explanation}
\vspace{-0.15cm}
\resizebox{\linewidth}{!}{
\begin{tabular}{lcc}
\toprule
\textbf{Criterion} & \textbf{Base} & \textbf{Ours} \\
\midrule
Visual support              & 2.27 & \textbf{3.94} \\
Anomaly-type correctness    & 2.03 & \textbf{3.75} \\
Avoids unsupported claims   & 1.91 & \textbf{3.84} \\
Decision justification      & 1.75 & \textbf{3.85} \\
\bottomrule
\end{tabular}
}
\vspace{-0.3cm}
\end{wraptable}
\noindent\textbf{Explanation Faithfulness.} To assess explanation quality independently of a VLM-based judge, we conduct a blinded human evaluation on 25 plots, comparing Base and \modelnamenc{} explanations for each plot. Six annotators rate the 50 anonymized explanations on visual support, anomaly-type correctness, avoidance of unsupported claims, and decision justification using a 1--5 scale while viewing the plot and ground-truth anomaly interval; model identity is hidden and explanation order is randomized. As shown in~\cref{tab:human_explanation}, \modelnamenc{} achieves higher mean ratings across all four criteria. Among 112 matched annotator--plot comparisons, \modelnamenc{} receives a higher overall score in 104 cases (92.9\%), while the Base model scores higher in 4 (3.6\%) and 4 are ties (3.6\%). The mean paired improvement is $+1.85$ points, with a 95\% confidence interval of $[1.66,\,2.05]$, providing human-evaluation evidence that \modelnamenc{} produces explanations that are more visually grounded, better aligned with the annotated anomaly type, and less likely to contain unsupported claims. See Appendix~\ref{apd:faithfulness} for full evaluation and paired-analysis details.

\noindent\textbf{Robustness to plot rendering.}
To evaluate the sensitivity of \modelnamenc{} to visualization choices, we conduct a controlled analysis on the test set while keeping the \modelnamenc{}-7B checkpoint, prompt, decoding procedure, and underlying time series fixed, and varying only the plot rendering. We consider lower image resolutions, different x-axis context, and expanded y-axis ranges. The results are reported in Table~\ref{tab:plot_robustness}. Overall, \modelnamenc{} remains robust to moderate rendering changes without retraining or adapting the model. Even when the image is compressed to $500\times175$ pixels, F$_1$ remains close to the original setting (73.56\% vs.\ 73.94\%). At the same time, rendering choices can substantially affect performance. In particular, providing 20\% local context around the anomaly improves F$_1$ by 7.93 points and overlap by 8.41 points. This suggests that effective visual resolution around the anomalous pattern is more important than image resolution alone. Performance is therefore expected to degrade when the rendering makes a local deviation difficult to distinguish, particularly for very long sequences or visually subtle anomalies.

\begin{table}[t]
\centering
\caption{\textbf{Sensitivity of \modelnamenc{}-7B to plot-rendering variations.} All settings use the same model checkpoint, prompt, decoding procedure, and underlying test time series.}
\vspace{0.1cm}
\label{tab:plot_robustness}
\resizebox{\linewidth}{!}{%
\begin{tabular}{lrrrrrrrr}
\toprule
Rendering variant & TP & FP & FN & Precision (\%) & Recall (\%) & F$_1$ (\%) & Overlap (\%) & $\Delta$F$_1$ (pts.) \\
\midrule
Original            & 576 & 223 & 183 & 72.09 & 75.89 & 73.94 & 25.35 & 0.00 \\
1000$\times$350     & 611 & 181 & 148 & 77.15 & 80.50 & 78.79 & 31.55 & +4.85 \\
500$\times$175      & 562 & 207 & 197 & 73.08 & 74.04 & 73.56 & 27.29 & $-$0.38 \\
Zoom context (20\%) & 630 & 150 & 129 & 80.77 & 83.00 & 81.87 & 33.76 & +7.93 \\
Y-range $1.5\times$ & 590 & 195 & 169 & 75.16 & 77.73 & 76.42 & 29.68 & +2.48 \\
Y-range $2\times$   & 574 & 213 & 185 & 72.94 & 75.63 & 74.26 & 29.15 & +0.32 \\
\bottomrule
\end{tabular}%
}
\vspace{-0.5cm}
\end{table}

%% file: assets/tables/tab_main_results.tex
\begin{table*}[t]
  \caption{\textbf{Anomaly Detection Performance on \datasetnc{}}.
  \colorbox{genVLM}{\phantom{\rule{1ex}{1ex}}} are General Large VLMs,
\colorbox{smVLM}{\phantom{\rule{1ex}{1ex}}} are Small VLMs, 
\colorbox{FM}{\phantom{\rule{1ex}{1ex}}} are Foundation Models, 
\colorbox{specLM}{\phantom{\rule{1ex}{1ex}}} are Specialized Large models, 
\colorbox{detector}{\phantom{\rule{1ex}{1ex}}} are Classical Anomaly Detectors,
\colorbox{ourVLM!10}{\phantom{\rule{1ex}{1ex}}} are \modelnamenc{} variants. Best performance is \textbf{{bold}}, and second best is \underline{underlined}. Green arrows (\textcolor{ForestGreen}{$\uparrow$}) indicate absolute improvement relative to the best baseline.}
\label{tab:anom_results}
   \vspace{-0.2cm}
  \centering

  {\fontsize{8pt}{11pt}\selectfont
   \resizebox{0.99\linewidth}{!}{%
     \begin{NiceTabular}{l lllllll}[colortbl-like]
       \toprule[1.2pt]
& \textbf{TP} & \textbf{FP} & \textbf{FN}
  & \textbf{Precision (\%)} & \textbf{Recall (\%)}
  & \textbf{F\textsubscript{1} (\%)}
  & \textbf{Overlap (\%)} \\

       \midrule[1.2pt]\midrule
    \rowcolor{genVLM}Grok-4-Fast ~\citep{grok2024}
    & 242 & \phantom{1}837 & 517
    & 22.43 & 31.88 & 26.33 & \phantom{1}3.26 \\
    \rowcolor{genVLM}LLaMA-4-Maverick (17B)~\citep{meta-llama4-2025}
    & 442 & 1128 & 317
    & 28.15 & 58.23 & 37.96 & 12.92 \\
       \midrule
    \rowcolor{smVLM}Qwen2.5-VL-7B~\citep{bai2025qwen2}
    & 517 & 1784 & 242
    & 22.47 & 68.12 & 33.79 & 14.04 \\
    \rowcolor{smVLM}Idefics3-8B~\citep{laurencconbuilding}
    & 123 & 2057 & 636
    & \phantom{1}5.64 & 16.21 & \phantom{1}8.37 & \phantom{1}2.25 \\

    \rowcolor{smVLM}SmolVLM-2.2B~\citep{marafioti2025smolvlm}
    & \phantom{1}49 & 1416 & 710
    & \phantom{1}3.34 & \phantom{1}6.46 & \phantom{1}4.41 & \phantom{1}1.29 \\
    \rowcolor{smVLM}LLaVA-7B~\citep{liu2023visual}
    & \phantom{11}9 & \phantom{1}523 & 750
    & \phantom{1}1.69 & \phantom{1}1.19 & \phantom{1}1.39 & \phantom{1}0.23 \\
       \midrule
    \rowcolor{FM}TimesFM (200M)~\citep{das2024decoder}& 196 & \phantom{1}543 & 563
    & 26.52 & 25.82 & 26.17 & \phantom{1}2.03 \\
    \rowcolor{FM}Chronos (120M)~\citep{ansarichronos}& 200 & \phantom{1}538 & 559
    & 27.10 &26.35 & 26.72 & \phantom{1}2.27 \\
    \midrule
    \rowcolor{specLM}AnomLLM (GPT-4o)~\citep{zhoucan}& 149 & \phantom{1}862 & 610
    & 14.74 & 19.63 & 16.84 & \phantom{1}4.04 \\
    \rowcolor{specLM}LLM-TSAD (GPT-4o)~\citep{parkdelving}& 477 & 5621& 282
    & \phantom{1}7.82 & 62.85 & 13.91 & 19.63 \\
    \rowcolor{specLM}LLMAD (GPT-4o)~\citep{liu2025large}& 349 & \phantom{1}402 & 410
    & 46.47 & 45.98 & 46.23 & 17.31 \\
    \rowcolor{specLM}VLM4TS (GPT-4o)~\citep{he2025harnessing}& 435 & \phantom{1}642 & 324
    & 40.39 & 57.31 & 47.39 & 17.94 \\

    \midrule
    \rowcolor{detector}Sub-PCA~\citep{aggarwal2016outlier}
    & 329 & \phantom{1}411 & 430
    & 44.46 & 43.35 & 43.90 & 17.41 \\
    \rowcolor{detector}Matrix Profile~\citep{linardi2020matrix}
    & 290 & \phantom{1}450 & 469
    & 39.19 & 38.21 & 38.69 & \phantom{1}8.33 \\
\rowcolor{detector}IForest~\citep{liu2008isolation}
    & 362 & \phantom{1}378 & 397
    & 48.92 & 47.69
    & 48.30 & 20.53 \\
       \midrule
    \rowcolor{ourVLM!20}\modelname{} (3B)
    & \underline{564}~\uparr~\textcolor{ForestGreen}{47} & \phantom{1}\underline{240}~\downarrg~\textcolor{ForestGreen}{138}  & \underline{195}~\downarrg~\textcolor{ForestGreen}{47}
    & \underline{70.15}~\comp{70.15}{48.92}  & \underline{74.31}~\comp{74.31}{68.12} 
    & \underline{72.17}~\comp{72.17}{48.30 }  & \textbf{27.07}~\comp{27.07}{20.53}  \\
    \rowcolor{ourVLM!20}\modelname{} (7B)
    & \textbf{576}~\uparr~\textcolor{ForestGreen}{59} & \phantom{1}\textbf{223}~\downarrg~\textcolor{ForestGreen}{155} & \textbf{183}~\downarrg~\textcolor{ForestGreen}{59}
    & \textbf{72.09}~\comp{72.09}{48.92}  & \textbf{75.89}~\comp{75.89}{68.12} 
    & \textbf{73.94}~\comp{73.94}{48.30}  & \underline{25.35}~\comp{25.35}{20.53}  \\
       \bottomrule
     \end{NiceTabular}
   }}
  \vspace{-0.3cm}
\end{table*}

%% file: assets/tables/TSB-U-AD.tex
\begin{table*}[t!]
  \centering
  \caption{\textbf{Anomaly Detection Performance on TSB-AD-U Benchmark}. \colorbox{genVLM}{\phantom{\rule{1ex}{1ex}}} are General Large VLMs,
\colorbox{smVLM}{\phantom{\rule{1ex}{1ex}}} are Small VLMs, 
\colorbox{FM}{\phantom{\rule{1ex}{1ex}}} are Foundation Models, 
\colorbox{specLM}{\phantom{\rule{1ex}{1ex}}} are Specialized Large models, 
\colorbox{detector}{\phantom{\rule{1ex}{1ex}}} are Classical Anomaly Detectors,
\colorbox{ourVLM!10}{\phantom{\rule{1ex}{1ex}}} are \modelnamenc{} variants. Best performance is \textbf{bold}, and second best is \underline{underlined}. Arrows indicate absolute changes relative to the best baseline.}
  \label{tab:tsbadu}
 \vspace{-0.2cm}
  {\fontsize{8pt}{11pt}\selectfont
  \resizebox{\linewidth}{!}{%
  \begin{NiceTabular}{l lll lll}[colortbl-like]
    \toprule[1.2pt]
    & \multicolumn{3}{c}{\textbf{Standard Metrics}} & \multicolumn{3}{c}{\textbf{Affiliation Metrics}} \\
    \cmidrule(lr){2-4} \cmidrule(lr){5-7}
    \textbf{Method} & \textbf{Precision (\%)} & \textbf{Recall (\%)} & \textbf{F\textsubscript{1} (\%)}
                    & \textbf{Precision (\%)} & \textbf{Recall (\%)} & \textbf{F\textsubscript{1} (\%)}
 \\
    \midrule[1.2pt]\midrule
     \rowcolor{genVLM}Grok-4-Fast ~\citep{grok2024} &27.95 & 15.53 & 16.78 & 32.43 & 33.24 & 32.20 \\
    \rowcolor{genVLM}LLaMA-4-Maverick (17B)~\citep{meta-llama4-2025} & 63.35 & 46.49 &48.89  & 76.07 & \underline{79.18} & 76.75 \\
    \midrule
    \rowcolor{smVLM}Qwen2.5-VL-7B~\citep{bai2025qwen2} & 66.18 &  48.85& 49.52 & 75.90 & 74.93 & 74.36 \\
    \rowcolor{smVLM}Idefics3-8B~\citep{laurencconbuilding} & 45.12 & 35.11 & 34.64 & 50.83 &59.02  &53.03  \\
    \rowcolor{smVLM}SmolVLM-2.2B~\citep{marafioti2025smolvlm} & 35.07 & 12.48 & 15.80 & 37.57 & 35.84 & 35.75 \\
    \rowcolor{smVLM}LLaVA-7B~\citep{liu2023visual} & 14.06 & \phantom{1}0.72 & \phantom{1}1.35 & 14.06 & \phantom{1}8.16 & 10.15 \\
    \midrule
    \rowcolor{FM}TimesFM (200M)~\citep{das2024decoder}
      & 32.32 & 43.21 & 32.19
      & 71.09 & 57.23 & 60.08 \\
    \rowcolor{FM}Chronos (120M)~\citep{ansarichronos}
      & 47.18 & \underline{55.37} & 47.23
      & 74.36 & 75.61 & 73.31 \\

    \midrule
    \rowcolor{specLM}AnomLLM (GPT-4o)~\citep{zhoucan}
      & 12.97 & 10.54 & 11.63
      & 13.05 & 7.08 & 8.42 \\

    \rowcolor{specLM}LLM-TSAD (GPT-4o)~\citep{parkdelving}
      & 31.74 & 28.40 & 29.98
      & 66.69 & 52.76 & 53.69 \\

    \rowcolor{specLM}LLMAD (GPT-4o)~\citep{liu2025large} & 32.96 & 37.54 & 22.91 & 67.50 & 73.37 & 64.16 \\
    \rowcolor{specLM}VLM4TS (GPT-4o)~\citep{he2025harnessing}
      & 29.78 & 40.10 & 19.84
      & 72.52 & 51.38 & 54.06 \\
    \midrule

    \rowcolor{detector}Sub-PCA~\citep{aggarwal2016outlier} &11.80 &16.82 &12.28 &48.22 &31.71& 34.99  \\
   
    \rowcolor{detector}Matrix Profile~\citep{linardi2020matrix} &9.40 &15.84 &9.91& 49.90 &41.44 &41.90 \\
    \rowcolor{detector}IForest~\citep{liu2008isolation} &12.44 &17.89 &13.05 &50.44 &41.51& 42.67  \\

    \midrule

    \rowcolor{ourVLM!20}\modelname{} (3B)& \textbf{77.93}\comp{77.93}{66.18 }  
      & 40.46~\comp{40.46}{55.37} 
      & \underline{49.59}~\comp{49.59}{49.52}
      & \underline{95.05}~\comp{95.05}{76.07  }
      & 71.69~\comp{71.69}{79.18}
      & \underline{78.36}~\comp{78.36}{76.75} \\

    \rowcolor{ourVLM!20}\modelname{} (7B)
      & \underline{75.75}~\comp{75.75}{66.18} 
      & \textbf{60.91}~\comp{60.91}{55.37} 
      & \textbf{62.91}~\comp{62.91}{49.52}
      & \textbf{95.27}~\comp{95.27}{76.07 }
      & \textbf{81.51}~\comp{81.51}{79.18}
      & \textbf{85.58}~\comp{85.58}{76.75} \\
    \bottomrule
  \end{NiceTabular}%
  }
  }
   \vspace{-0.6cm}
\end{table*}

%% file: assets/figures/qual_visanom.tex
\begin{figure*}[!t]
    \centering    
    \includegraphics[width=0.99\linewidth]{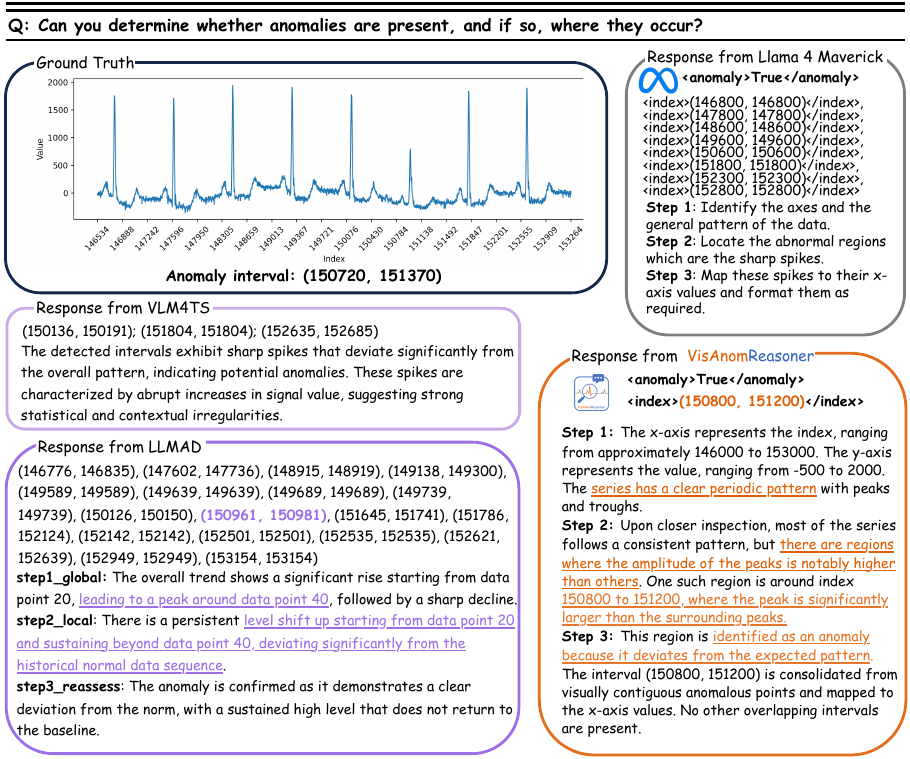}
\vspace{-0.2cm}
\caption{\textbf{Qualitative Examples of Anomaly Reasoning.}
\modelname{} precisely localizes the anomalous interval with visually grounded, structured reasoning, whereas other methods exhibit coarser localization or produce numerous spurious intervals with less grounded explanations.}
    \label{fig:qual}
   \vspace{-0.3cm}
\end{figure*}

%% file: assets/figures/base_vs_SFT.tex
\begin{wrapfigure}{r}{0.52\textwidth}
\vspace{-0.45cm}
\centering
\includegraphics[width=\linewidth]{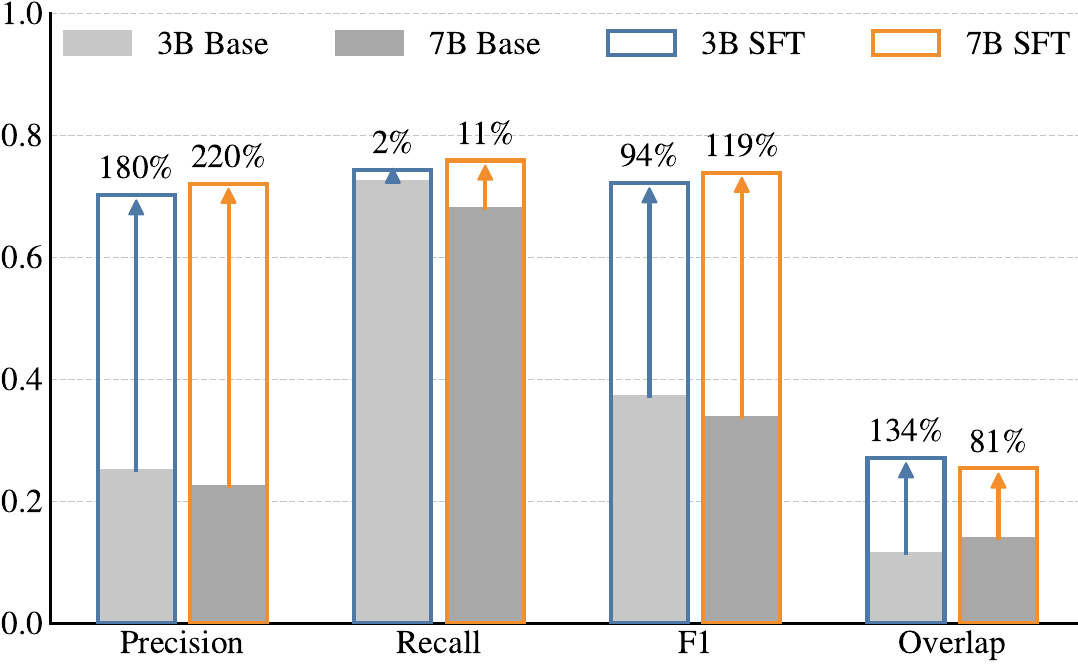}
\vspace{-0.65cm}
\caption{\textbf{Comparison of base Qwen2.5-VL models and their variants trained using supervised fine-tuning} (\modelname{}).
Arrows denote relative improvements due to supervised fine-tuning.}
\label{fig:base_vs_SFT}
\vspace{-0.3cm}
\end{wrapfigure}

%% file: sections/05_conclusion.tex
\section{Conclusion}\label{sec:conclusion}
This work introduces vision-language reasoning over time-series data, where models must align temporal evidence, anomaly intervals, and natural-language explanations in a single prediction.
To address the lack of suitable supervision for VLM-based anomaly detection, we  curate
\dataset{}, an explanation-augmented benchmark that enables supervised learning of plot-grounded anomaly reasoning, and develop \modelname{}, a compact VLM trained to jointly localize anomaly intervals and generate visual-temporal explanations. 
Across \datasetnc{} and \textsc{TSB-AD-U}, \modelnamenc{} consistently outperforms general-purpose VLMs, specialized LLM/VLM anomaly detectors, time-series foundation models, and classical methods, with large gains in precision, F\textsubscript{1}, and temporal alignment. 
Our analyses show that reasoning supervision outperforms interval-only fine-tuning, while multi-model supervision outperforms single-model supervision. Both VLM and human evaluations show that \modelnamenc{} produces higher-quality, more visually grounded explanations than the Base model, while rendering perturbations demonstrate robustness to moderate visualization changes.
These results highlight explanation-driven supervision as a practical path toward interpretable, reasoning-centric time-series anomaly detection.

%% file: sections/06_appendix.tex
\newpage
\appendix
\onecolumn

\input{assets/appendix/benchmark}

\input{assets/appendix/models}

\input{assets/appendix/implementation_details}

\input{assets/appendix/qual_example2}
\input{assets/appendix/effect_anomaly_free_supervision}

\input{assets/appendix/explanation_faithfulness}
\input{assets/appendix/statistical_support}

\input{assets/appendix/additional_results}

\input{assets/appendix/plot_render_robustness}

\input{assets/appendix/efficiency}

\input{assets/appendix/limitations}
     

%% file: assets/appendix/benchmark.tex
\section{Benchmark Details}\label{apd:benchmark}

\subsection{\datasetnc{}}\label{apd:VisAnomBench_constrcut}

\datasetnc{} is constructed from four public time-series anomaly detection benchmarks:
KPI~\citep{xu2018unsupervised}, GutenTAG~\citep{WenigEtAl2022TimeEval}, UCR-EGI~\citep{gao2020ensemble}, and the UCR Time Series Anomaly Datasets~\citep{wu2021current}.
These benchmarks span diverse application domains and cover a wide range of anomaly types, sequence lengths, temporal patterns, and data characteristics.

To ensure consistency and data quality across sources, a unified preprocessing procedure is applied.
Each original time series is segmented to satisfy the following criteria:
\begin{itemize}[itemsep=0.2ex, topsep=0pt]
    \item The anomaly ratio does not exceed 10\%, with a target range of $(0.01, 0.10)$.
    \item The anomaly interval is centered between 30\% and 70\% of the segment length.
    \item The segment length is at least 200 time steps.
\end{itemize}

To mitigate the impact of known anomaly mislabeling issues in public datasets~\citep{wu2021current}, we apply an additional verification step using multiple general-purpose large VLMs.
Specifically, each candidate time-series segment is rendered as a plot and provided to four independent large VLMs together with the ground-truth anomaly interval indices.
The prompt template for obtaining anomaly explanations is shown in~\cref{fig:ts_anomaly_prompt}.
Segments for which the majority of models either fail to identify any anomaly or consistently predict substantially different anomaly intervals are excluded from \datasetnc{}.
This verification removes only 5 of 3,321 candidate samples (0.15\%), limiting its effect on the benchmark composition. Manual inspection of these cases indicates ambiguous or visually inconsistent annotations rather than simply difficult anomaly patterns. To further assess whether this filtering materially affects the reported results, we add all five excluded samples back to the held-out evaluation. The precision, recall, and F\textsubscript{1} of \modelnamenc{}-7B change from 72.09\%, 75.89\%, and 73.94\% to 72.24\%, 72.88\%, and 72.56\%, respectively, corresponding to a 1.38 percentage point decrease in F\textsubscript{1}. Thus, the filtering step removes only a negligible fraction of the candidate samples, and the main performance conclusions remain unchanged when the excluded samples are restored.

After filtering, \datasetnc{} contains 2,576 training time series and 740 test time series, with sequence lengths ranging from 200 to over 100K points. As summarized in~\cref{tab:dataset_split_stats}, the training and test splits have mean sequence lengths of 6,882 and 6,259 points, mean anomaly lengths of 305 and 261 points, and mean anomaly ratios of 5.0\% and 4.5\%, respectively.

\input{assets/appendix/prompt_template}

\begin{table}[!htbp]
\centering
\small
\setlength{\tabcolsep}{7pt}
\renewcommand{\arraystretch}{1.1}
\caption{\textbf{\datasetnc{} Statistics.}
The train and test splits contain univariate time series with lengths ranging from 200 points to over 100K points.}
\label{tab:dataset_split_stats}
\begin{tabular}{lcccccc}
\toprule
\textbf{Split} & \textbf{\# TS} & \textbf{Min Len.} & \textbf{Max Len.} &
\textbf{Mean Len.} & \textbf{Avg. Anom. Len.} & \textbf{Avg. Anom. Ratio} \\
\midrule
Train & 2576 & 200 & 105472 & 6882 & 305 & 5.0\% \\
Test  & 740  & 200 & 102400 & 6259 & 261 & 4.5\% \\
\bottomrule
\end{tabular}
\end{table}

\subsubsection{Candidate Generation and Reward-Based Selection}\label{apd:rating}

For each time-series segment, we construct a pool of candidate structured outputs using four general-purpose VLMs: Grok-4-Fast~\citep{grok2024}, LLaMA-4-Maverick~\citep{meta-llama4-2025}, Gemma-3-27B-IT~\citep{gemma_2025}, and Qwen2.5-VL-32B ~\citep{bai2025qwen2}.
Each candidate includes an anomaly decision, predicted anomaly interval(s), and a natural-language reasoning trace.
The candidates are ranked according to the composite reward defined in Eq.~\eqref{eq:reward}, and the highest-scoring candidate is retained as the supervision target for the corresponding segment. The composite reward consists of an anomaly localization term and three explanation-quality terms.
The anomaly localization score is computed from the predicted and ground-truth anomaly intervals.
The explanation-quality scores are obtained using Qwen2.5-VL-72B as a judge.
Given the plot and generated explanation, the judge assigns scores in $[0,1]$ for visual groundedness, axis awareness, and clarity.
The weighting coefficients are set to $\lambda_{\text{ano}}=0.3$, $\lambda_{\text{vis}}=0.3$, $\lambda_{\text{axi}}=0.1$, and $\lambda_{\text{cla}}=0.3$.
This reward-based selection procedure determines the final supervision source for each training series.
As shown in~\cref{tab:dataset_source_composition}, the selected supervision comes from all four generator models, with LLaMA-4-Maverick contributing the largest share.
The selected outputs are subsequently used for standard supervised fine-tuning of \modelnamenc{}.
The judge prompt template is presented in~\cref{fig:ts_anomaly_judge_prompt}.

\input{assets/appendix/judge_template}

\begin{table}[!htbp]
\centering
\small
\setlength{\tabcolsep}{6pt}
\caption{\textbf{Composition of Reward-Selected Supervision in \datasetnc{}.}
Counts and percentages indicate which generator model produced the selected supervision target for each training series.}
\label{tab:dataset_source_composition}
\begin{tabular}{lcccc}
\toprule
 & \textbf{Grok-4-Fast} & \textbf{LLaMA-4-Maverick} & \textbf{Gemma-3-27B-IT} & \textbf{Qwen2.5-VL-32B} \\
\midrule
Count   & 605  & 1007 & 549 & 415 \\
Percent & 23.5\% & 39.1\% & 21.3\% & 16.1\% \\
\bottomrule
\end{tabular}
\end{table}

\subsubsection{Explanation Quality Evaluation}

Human validation at the full scale of \datasetnc{} is impractical because the benchmark contains thousands of time series from multiple domains, which would require domain experts for reliable verification.
We manually inspected a random subset of selected and rejected candidates during construction and found that selected explanations were generally more visually grounded, more plot-specific, and more consistent with the labeled anomaly intervals than lower-scoring alternatives.
Representative examples are shown in~\cref{fig:selected_discarded_examples}.

\input{assets/appendix/think_examples}

\subsection{\textsc{TSB-AD-U}}

We use the \textsc{TSB-AD-U} benchmark evaluation subset for univariate time-series anomaly detection as in existing works ~\citep{parkdelving, zhoucan}. The \textsc{TSB-AD} benchmark originally collected 13 univariate and 20 multivariate datasets and, after curation, expanded to 23 univariate and 17 multivariate datasets. Following the TSB-AD evaluation protocol, we select the same eight univariate datasets, listed in~\cref{tab:tsbad_u_stats}, for fair comparison with existing models. All experiments on \textsc{TSB-AD-U} in this paper are conducted on this fixed evaluation set.

\begin{table}[!htbp]
\centering
\small
\setlength{\tabcolsep}{6pt}
\renewcommand{\arraystretch}{1.1}
\caption{\textbf{TSB-AD-U Evaluation Benchmark Dataset Statistics.}
The total length is computed as the product of the average time-series length and the number of time series (Count).}
\label{tab:tsbad_u_stats}
\resizebox{\linewidth}{!}{
\begin{tabular}{lccccccc}
\toprule
\textbf{Dataset} & \textbf{Count} & \textbf{Dim.} & \textbf{Total Len.} &
\textbf{Avg. \# Anom.} & \textbf{Avg. Anom. Len.} & \textbf{Anom. Ratio} & \textbf{Category} \\
\midrule

NEK~\citep{si2024timeseriesbench}      & 8  & 1 & \textbf{8,584}  & 2.9  & 51.1  & 8.0\% & P\&Seq \\

TAO~\citep{noaa_tao_2025}      & 2  & 1 & \textbf{20,000} & 838.7 & 1.1   & 9.4\% & P\&Seq \\

MSL~\citep{hundman2018detecting}      & 7  & 1 & \textbf{23,111} & 1.3  & 130.0 & 5.8\% & Seq \\

Power~\citep{keogh2007finding}    & 1  & 1 & \textbf{35,040} & 4.0  & 750.0 & 8.5\% & Seq \\

Daphnet~\citep{bachlin2009wearable}  & 1  & 1 & \textbf{38,774} & 6.0  & 384.3 & 5.9\% & Seq \\

YAHOO~\citep{laptev_s5_2015}    & 30 & 1 & \textbf{45,270} & 5.5  & 2.5   & 0.6\% & P\&Seq \\

SED~\citep{boniol2021sand}      & 2  & 1 & \textbf{59,998} & 14.7 & 64.0  & 4.1\% & Seq \\

TODS~\citep{lai2021revisiting}     & 13 & 1 & \textbf{65,000} & 97.3 & 18.7  & 6.3\% & P\&Seq \\
\bottomrule
\end{tabular}
}
\end{table}

%% file: assets/appendix/prompt_template.tex
\begin{figure}[t] 
\centering
\resizebox{0.95\textwidth}{!}{
\begin{planbox}{Prompt Template for Eliciting Time-Series Anomaly Reasoning Traces} 
\small

You are given a \textbf{time-series plot image} (not raw values) and a short context: \texttt{\{ts\_context\}}\\

Your task is to (i) decide whether anomalies are present, (ii) localize them as inclusive
index/timestamp intervals exactly as shown on the plot axes, and (iii) provide concise,
step-by-step reasoning grounded in what is visible.
If citing numbers, estimate \textbf{only} from axis ticks or labels.

Please follow these steps:

\begin{itemize} \itemsep0em
\item Give a short image description relevant to anomaly detection, focusing on axes type,
visible trend or seasonality, and any obvious spikes, drops, or level shifts.

\item Provide a brief rationale distinguishing normal behavior (baseline, seasonality,
variance) from visible deviations. Numeric estimates should be derived only from axis labels
or tick spacing and noted approximately.

\item Generate a simplified \textbf{step-by-step reasoning process} consisting of 3--4 numbered steps.
Each step must be specific, non-redundant, and grounded in visual evidence.

\item Present the \textbf{final decision} using the \textbf{STRICT schema} below.
\end{itemize}

\textbf{STRICT schema and rules:}
\begin{itemize} \itemsep0em
\item Intervals are \textbf{inclusive} of endpoints.
\item Use x-axis values \textbf{as shown} (timestamps or integer indices); do not invent precision.
\item Merge visually contiguous anomalous points into a single interval; intervals must be sorted
and non-overlapping.
\item When anomalies exist, wrap \textbf{each} $(\text{start}, \text{end})$ pair in its own
\texttt{\textless index\textgreater...\textless/index\textgreater} tag.
\item When no anomalies exist, output only
\texttt{\textless anomaly\textgreater False\textless/anomaly\textgreater}
with no \texttt{\textless index\textgreater} tags.
\end{itemize}

\textbf{Sections to include before the two final lines (use these exact headings):}
\begin{itemize} \itemsep0em
\item \texttt{Image Description:}
\begin{itemize}
    \item  (1–3 sentences about the plot relevant to anomalies.)
\end{itemize}
\item \texttt{Rationales:}
\begin{itemize}
    \item  (Concise bullet 1: expected behavior from context.)
    \item (Concise bullet 2: visible deviations with axis-aware estimates when needed.) 
    \item (Optional concise bullet 3.) 
\end{itemize}
\item \texttt{Let's think step by step.}
\begin{itemize}
    \item  Step 1: (Identify axes scale and baseline; note any seasonality.) 
    \item Step 2: (Locate candidate abnormal regions and justify deviations.) 
    \item Step 3: (Consolidate adjacent points into inclusive intervals; map to x-axis values; ensure sorted, non-overlapping.)
\end{itemize}

\item \texttt{The anomaly is:}
\begin{itemize}
    \item  \{gt\_anomaly\_intervals\}
\end{itemize}

\end{itemize}

\textbf{Final parseable output (after these sections, produce exactly these two lines, then stop):}
\begin{itemize} \itemsep0em
\item \texttt{\textless anomaly\textgreater True/False\textless/anomaly\textgreater
\textless index\textgreater(start\_1,end\_1)\textless/index\textgreater,
\textless index\textgreater(start\_2,end\_2)\textless/index\textgreater}
\item \texttt{\textless think\textgreater Step 1: ... Step 2: ... Step 3: ...
\textless/think\textgreater}
\end{itemize}

\end{planbox}
}
\vspace{-0.1cm}
\caption{\textbf{Prompt for Time-Series Anomaly Reasoning.}
The prompt enforces axis-aware localization of predicted anomaly interval(s) and concise, visually grounded step-by-step explanations under a strict, parsable output schema.}
\label{fig:ts_anomaly_prompt}
\end{figure}

%% file: assets/appendix/judge_template.tex
\begin{figure}[t]
\centering
\resizebox{0.95\textwidth}{!}{
\begin{planbox}{Prompt Template for Rating Time-Series Anomaly Reasoning Traces}
\small

You are evaluating a model's reasoning for time-series anomaly detection. Use the attached image and the inputs below. 

Inputs:\\
- Context: \texttt{\{context\}}\\
- Decision (Line 1): \texttt{\{decision\}}\\
- Reasoning (inside \texttt{\textless think\textgreater}): \texttt{\{reasoning\}}\\

Evaluate along three dimensions:\\
1. Visual Groundedness: How well does the reasoning reference visible patterns (spikes, drops, shifts) in the plot?\\
2. Axis Awareness: Are timestamps/indices and value ranges consistent with the actual axes? Penalize hallucinated precision.\\
3. Clarity: Is the reasoning logically ordered, non-redundant, and clearly connected from observation to conclusion?\\

Return your scores as three numbers in this exact format:\\
\texttt{VISUAL: \textless number in [0,1]\textgreater}\\
\texttt{AXIS: \textless number in [0,1]\textgreater}\\
\texttt{CLARITY: \textless number in [0,1]\textgreater}\\

Do NOT add any other text, explanation, or formatting. Only those three lines.

\end{planbox}
}
\vspace{-0.1cm}
\caption{\textbf{Judge Prompt for Rating Time-Series Anomaly reasoning.}
The prompt scores model-generated reasoning along visual groundedness, axis awareness, and clarity using the time series plot.}
\vspace{-0.4cm}
\label{fig:ts_anomaly_judge_prompt}
\end{figure}

%% file: assets/appendix/think_examples.tex
\begin{figure}[t]
\centering
\begin{tcolorbox}[
  colback=gray!5,
  colframe=black!20,
  boxrule=0.5pt,
  width=0.99\textwidth,
  left=3pt,
  right=3pt,
  top=4pt,
  bottom=4pt
]
\small

\vspace{0.4em}
\textbf{Preferred examples contain more detailed, axis-aware justifications for the selected intervals, including the observed scale of the plot, the surrounding baseline behavior, and the visual deviation that supports the anomaly label.}

\begin{examplebox}
<anomaly>True</anomaly><index>(2017-01-09 05:41:00, 2017-01-09 05:44:00)</index>

<think>Step 1: X-axis uses datetime ticks (e.g., 2017-01-09) suggesting minute-level resolution within days; y-axis scales linearly from 0 to 0.25; baseline hovers at ~0.05-0.10 with random small oscillations, no clear seasonal cycles. Step 2: Candidate anomalous region is the cluster of high points around 2017-01-09 05:41:00 to 05:44:00, where the line spikes to the top of the y-range (~0.25), deviating sharply from the surrounding low-variance noise. Step 3: This forms a single contiguous interval (2017-01-09 05:41:00, 2017-01-09 05:44:00); no other clusters qualify; sorted ascending with no overlaps.</think>
\end{examplebox}

\begin{examplebox}
<anomaly>True</anomaly><index>(2016-07-26 00:55:00, 2016-07-26 01:05:00)</index>

<think>Step 1: The x-axis uses datetime scale with ticks at roughly hourly intervals from 2016-07-25 23:00 to 2016-07-26 01:00; y-axis spans 0 to 2. Step 2: A candidate anomaly appears as a sharp upward spike peaking above 1.5 near 2016-07-26 01:00, violating the expected steady behavior by representing a sudden load increase not seen elsewhere in the visible range. ... </think>
\end{examplebox}

\vspace{0.3em}
\textbf{Discarded examples provide weaker support for the selected intervals: they tend to describe anomalies only at a high level, give limited evidence for why the intervals are abnormal, or select intervals that are not well aligned with the most clearly described visual deviations.}

\begin{examplebox2}
<anomaly>True</anomaly><index>(2017-09-04 21:00:00, 2017-09-04 21:00:00)</index>, <index>(2017-09-06 03:00:00, 2017-09-06 03:00:00)</index>, ...

<think>Step 1: Identified the axes scales and baseline value for metric 'g'. Step 2: Located five distinct spikes drastically differing from the baseline. Step 3: Formatted each spike location into an inclusive index pair and sorted them to ensure accuracy.</think>
\end{examplebox2}

\begin{examplebox2}
<anomaly>True</anomaly><index>(2016-07-26 01:45:00, 2016-07-26 02:15:00)</index>

<think>Step 1: Identify the axes and baseline; the plot shows CPU utilization with a downward trend followed by an upward trend. Step 2: Locate candidate abnormal regions; the dip around 2016-07-25 20:00:00 and the spike around 2016-07-26 02:00:00 are notable. Step 3: Consolidate adjacent points into inclusive intervals; the spike around 2016-07-26 02:00:00 is an anomaly.</think>
\end{examplebox2}
\end{tcolorbox}
\vspace{-0.4cm}
\caption{\textbf{Representative Selected and Discarded Explanations.}
Preferred explanations provide axis-aware and visually grounded support for the predicted anomaly interval, while discarded examples are more generic or weakly supported by the plot.}
\vspace{-0.5cm}
\label{fig:selected_discarded_examples}
\end{figure}

%% file: assets/appendix/models.tex
\section{Model Details}\label{apd:models}

We compare against foundation models, specialized large models, and classical detectors.

\noindent \tightcolorbox{FM}{\textbf{Foundation Models.}}
\begin{itemize}[itemsep=0.4ex, parsep=0pt, topsep=0pt, leftmargin=0.4cm]
    \item Chronos~\citep{ansarichronos} converts continuous time-series values into discrete tokens via scaling and quantization, then trains a T5-based model with a standard cross-entropy objective for anomaly detection.
    \item TimesFM~\citep{das2024decoder} pretrains a decoder-only attention model on large-scale time-series data using patch-based inputs, learning general representations applicable to forecasting and anomaly detection.
\end{itemize}

\noindent \tightcolorbox{specLM}{\textbf{Specialized Large Models.}} 
\begin{itemize}[itemsep=0.4ex, parsep=0pt, topsep=0pt, leftmargin=0.4cm]
    \item  AnomLLM~\citep{zhoucan} performs time-series anomaly detection by converting time series into textual representations and prompting large language models such as GPT-4 and Qwen with 21 prompt templates to identify anomalous segments.
   \item LLM-TSAD~\citep{parkdelving} extends AnomLLM by explicitly incorporating time-series indices into the prompt and providing trend and residual decompositions of the time series as visual inputs.
   \item LLMAD~\citep{liu2025large} relies on carefully structured prompts that encode domain rules, explicit anomaly type definitions, rarity constraints, and step-by-step reasoning instructions, combined with in-context examples to guide LLMs in detecting and explaining anomalies from time-series data.
    \item VLM4TS~\citep{he2025harnessing} segments raw time series into windows and converts them into images for multi-scale feature extraction, producing candidate anomalous intervals; a vision–language model (\eg GPT-4o) is then prompted to incorporate global temporal context to refine the detections.
\end{itemize}

\noindent \tightcolorbox{detector}{\textbf{Classical Detectors.}}
\begin{itemize}[itemsep=0.4ex, parsep=0pt, topsep=0pt, leftmargin=0.4cm]
    \item \textbf{Isolation Forest (IForest)}~\citep{liu2008isolation} detects anomalies by recursively partitioning the data space using random splits. Samples that can be isolated with fewer splits, corresponding to shorter path lengths in the resulting binary trees, are assigned higher anomaly scores.

    \item \textbf{Sub-PCA}~\citep{aggarwal2016outlier} applies principal component analysis to local subsequences by projecting them into a low-dimensional linear subspace. Subsequences that exhibit large reconstruction errors under this projection are identified as anomalies, reflecting violations of the assumed linear structure.

    \item \textbf{Matrix Profile}~\citep{linardi2020matrix} identifies anomalies by measuring the distance between each subsequence and its nearest neighbor within the time series. Subsections with unusually large nearest-neighbor distances are flagged as anomalous, indicating deviation from dominant temporal behavior.
\end{itemize}

%% file: assets/appendix/implementation_details.tex
\section{Evaluation Metrics}\label{apd:imple}
We evaluate anomaly detection performance using interval-level metrics, including
True Positives, False Positives, False Negatives, Precision, Recall, F\textsubscript{1},
and the Overlap score. We also report Precision, Recall, and F\textsubscript{1} obtained via the point-wise evaluation and the affiliation-based evaluation, where the latter accounts for temporal proximity between predicted and ground-truth anomaly intervals.

\noindent \textbf{True Positives, False Positives, and False Negatives.}
Following~\citep{baireddy2021spacecraft}, let
$\mathcal{A}^\star=\{(s_i^\star,e_i^\star)\}_{i=1}^{m}$, $1 \le s_i^\star < e_i^\star \le T$ denote the set of ground-truth anomaly intervals, and $A=\{(s_i,e_i)\}_{i=1}^{\hat m}$, $1 \le s_i < e_i \le T$
denote the set of predicted anomaly intervals.

\begin{itemize}[itemsep=0.8ex, parsep=0pt, topsep=0pt, leftmargin=0.7cm]
    \item \textbf{True Positives.} A predicted interval $(s_j,e_j)$ is counted as a true positive (TP) if it overlaps with a ground-truth interval in $\mathcal{A}^\star$ (if multiple predicted intervals overlap with the same ground-truth interval, only one TP is counted).

    \item \textbf{False Positives.} A predicted interval $(s_j,e_j)$ is counted as a false positive (FP) if it does not overlap with any ground-truth interval in $\mathcal{A}^\star$.

    \item \textbf{False Negatives.} A ground-truth interval  $(s_j^\star,e_j^\star)$ is counted as a false negative (FN)
    if it does not overlap with any predicted interval in \(A\).
\end{itemize}

\noindent For methods that take a time-series plot as input, including \modelnamenc{}, each predicted interval $(s_j,e_j)$ is extended by 100 time steps on each side for VisAnomBench and by 30 time steps for TSB-U-AD before computing all metrics.

\noindent \textbf{Interval-Based Precision, Recall, and F\textsubscript{1}.}
Using the definitions above for TPs, FPs, and FNs, the Precision, Recall, and F\textsubscript{1} metrics defined at the interval level are:
\begin{equation}
\mathrm{Precision} = \frac{N_\mathrm{TP}}{N_\mathrm{TP} + N_\mathrm{FP}} , \quad
\mathrm{Recall} = \frac{N_\mathrm{TP}}{N_\mathrm{TP} + N_\mathrm{FN}} , \quad
\mathrm{F}_1 = 2\frac{\mathrm{Precision} \cdot \mathrm{Recall}}{\mathrm{Precision} + \mathrm{Recall}} ,
\end{equation}
where $N_\mathrm{TP}$, $N_\mathrm{FP}$, and $N_\mathrm{FN}$ are the counts of all TPs, FPs, and FNs, respectively.

\noindent \textbf{Overlap Score.}
We report the Overlap score, adapted from the one introduced in~\citep{gao2020ensemble}, which measures temporal alignment
between predicted and ground-truth anomalies at the point level.
Let $\mathcal{T}_{\mathcal{A}^\star}$ denote the set of all time points in $\mathcal{A}^\star$ and $\mathcal{T}_A$ the set of all time points in $A$.
The Overlap score is defined as
\begin{equation}
\label{eq:overlap}
\mathrm{Overlap} =
\frac{\lvert \mathcal{T}_{\mathcal{A}^\star} \cap \mathcal{T}_A \rvert}
{\max\bigl(\lvert \mathcal{T}_{\mathcal{A}^\star} \rvert, \lvert \mathcal{T}_A \rvert\bigr)}.
\end{equation}
The normalization in Eq. \eqref{eq:overlap} discourages trivial solutions that predict excessively long anomaly intervals.
The score ranges from $0$ (no overlap) to $1$ (perfect temporal alignment).

\noindent \textbf{Standard Precision, Recall, and F\textsubscript{1}.}
For standard point-wise evaluation, the anomaly detection task is cast as a binary classification problem at the timestamp level.
Each time point is labeled as either anomalous (1) or normal (0), and Precision, Recall, and F\textsubscript{1}  are computed using the
conventional definitions for binary classification.

\noindent \textbf{Affiliation Precision, Recall, and F\textsubscript{1}.}
Affiliation metrics compute event-level Precision, Recall, and F\textsubscript{1} based on the temporal distance between predicted and ground-truth anomalies~\citep{huet2022local}, providing robustness to small misalignments while evaluating correct event affiliation.

%% file: assets/appendix/qual_example2.tex
\section{Additional Qualitative Examples}\label{apd:qual_ex}

\begin{figure*}[!t]
    \centering    
    \includegraphics[width=0.99\linewidth]{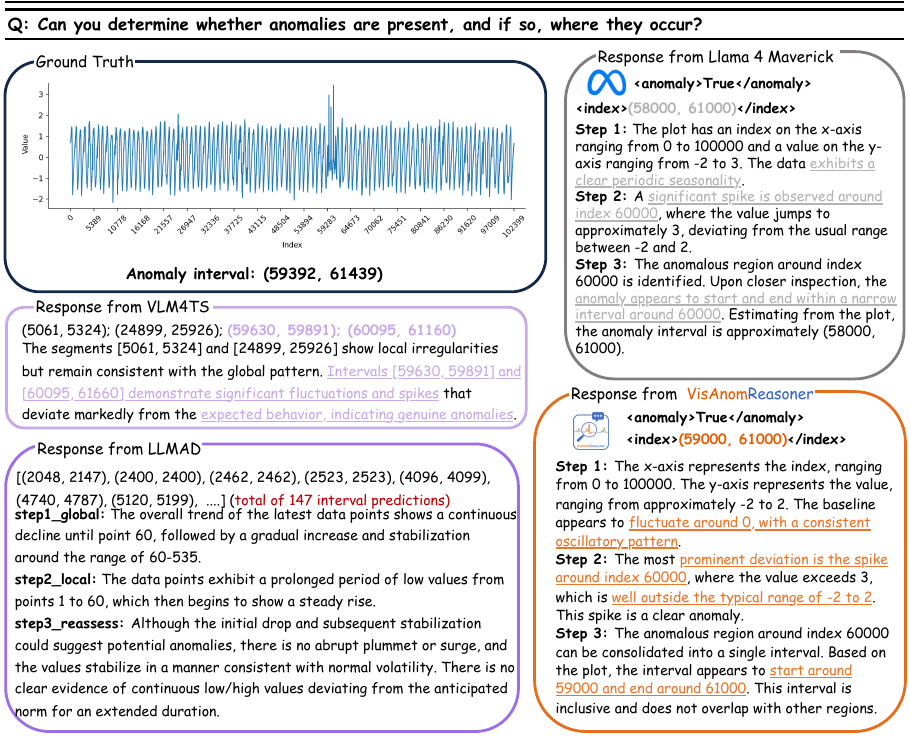}
    \vspace{-0.1cm}
\caption{\textbf{Additional Qualitative Examples of Anomaly Detection and Reasoning Across Models.}
\modelname{} precisely localizes the anomalous interval with visually grounded, structured reasoning; LLaMA~4~Maverick~\cite{meta-llama4-2025} correctly identifies the anomaly but exhibits lower localization accuracy, while the other methods continue to overflag anomaly intervals.}
    \label{fig:qual_2}
\end{figure*}

\cref{fig:qual_2} presents additional qualitative examples.
VLM4TS~\citep{he2025harnessing} continues to flag normal regions as anomalous; however, its performance on this example improves, as two of the predicted intervals together cover about two-thirds of the ground-truth anomaly region. The accompanying explanation partially reflects the trend in the plot, but remains shallow and generic. LLMAD~\citep{liu2025large} is strongly affected by normal fluctuations in the time series, resulting in 147 predicted anomaly intervals. Its explanation remains weakly grounded and does not correspond to the observed temporal behavior.
LLaMA~4~Maverick~\cite{meta-llama4-2025} produces a largely correct anomaly prediction and provides a coherent explanation. However, compared to \modelnamenc{}, its localization is less precise, as it predicts the anomaly onset approximately 1{,}392 time steps earlier than the ground truth.

%% file: assets/appendix/effect_anomaly_free_supervision.tex
\section{Additional Analysis}\label{apd:analysis}

\subsection{Effect of Anomaly-Free Supervision}\label{apd:normal_only}

The original \datasetnc{} contains anomaly-positive examples and is designed primarily for anomaly localization and explanation. Although predicting an anomaly interval implicitly identifies the remaining regions as non-anomalous, the SFT training set contains no fully anomaly-free examples. We therefore conduct an additional experiment to evaluate whether explicit negative supervision improves robustness to normal inputs and affects localization on anomaly-positive examples.

We construct anomaly-free examples from GutenTAG, where anomalies are injected at known locations. Using the official train/test split, we extract one fixed-length segment before the first annotated anomaly from each eligible time series and verify that it does not overlap any anomalous interval. This produces 413 anomaly-free training examples and 247 held-out test examples. No samples are selected based on model predictions, VLM judgments, or manual inspection. The anomaly-free and anomaly-positive examples use the same plotting pipeline, x-axis convention, rendering settings, and prompt format. 
For anomaly-free examples, the target contains \texttt{<anomaly>False</anomaly>} and no \texttt{<index>} prediction.
We fine-tune \modelnamenc{} using the same model, optimization setup, prompt format, and anomaly-positive training data as the original experiment, changing only the addition of these anomaly-free examples. The resulting model correctly predicts all 247/247 held-out anomaly-free samples as non-anomalous under this controlled evaluation. It also improves performance on the original 740 anomaly-positive \datasetnc{} test series, as shown in Table~\ref{tab:normal_only}.

\begin{table}[t!]
\centering
\caption{\textbf{Effect of adding anomaly-free examples during SFT.} Both models are evaluated on the same anomaly-positive \datasetnc{} test set.}
\label{tab:normal_only}
\resizebox{\linewidth}{!}{%
\begin{tabular}{lrrrrrrr}
\toprule
Model & TP & FP & FN & Precision (\%) & Recall (\%) & F$_1$ (\%) & Overlap (\%) \\
\midrule
\modelname{} 
& 576 & 223 & 183 & 72.09 & 75.89 & 73.94 & 25.35 \\
\modelname{} + anomaly-free SFT
& 618 & 138 & 141 & \textbf{81.75} & \textbf{81.42} & \textbf{81.58} & \textbf{33.17} \\
\bottomrule
\end{tabular}%
}
\vspace{-0.2cm}
\end{table}

Adding anomaly-free supervision reduces false positives from 223 to 138 while also reducing false negatives from 183 to 141. To understand why negative supervision also improves localization, we compare predictions from the two models on each of the same 740 anomaly-positive test series, containing 759 ground-truth anomaly intervals. The TP/FP/FN outcomes are unchanged for 589 series. Among the changed cases, 80 series exhibit a particularly informative correction: the original model predicts an unsupported interval while missing the ground-truth anomaly, whereas the model trained with anomaly-free examples shifts its prediction to the annotated anomalous region. Across these 80 series, this recovers 80 missed anomalies and removes 83 false-positive intervals. In another 22 series, the extended model removes 42 false-positive intervals without reducing recall. The remaining 49 series exhibit mixed changes, giving an overall difference of $+42$ TP, $-85$ FP, and $-42$ FN. Inspection of the corrected cases shows that many removed false positives correspond to benign periodic peaks, repeated motifs, or local fluctuations outside the annotated anomaly interval. These results suggest that anomaly-free supervision provides useful negative contrast: rather than only improving rejection of fully normal inputs, it helps the model distinguish benign visual patterns from the true anomalous region, reducing unsupported predictions and improving interval localization.

%% file: assets/appendix/explanation_faithfulness.tex
\subsection{Explanation Faithfulness}\label{apd:faithfulness}

To evaluate explanation quality independently of a VLM-based judge, we conduct a blinded human evaluation comparing explanations generated by the Base model and \modelnamenc{} on the same held-out test plots. We select 25 plots and obtain one explanation from each model, yielding 50 anonymized explanation instances. Six volunteer annotators independently evaluate the explanations while viewing the corresponding plot and ground-truth anomaly interval. Model identity is hidden and explanation order is randomized.
Annotators rate each explanation on four criteria using a 1--5 scale: (1) visual support, (2) anomaly-type correctness, (3) avoidance of unsupported claims, and (4) decision justification. As shown in~\cref{tab:human_explanation}, \modelnamenc{} receives higher mean ratings across all four criteria.\looseness-1
 
For paired analysis, we use only cases in which the same annotator rated both the Base and \modelnamenc{} explanations for the same plot. Because not all annotators completed every item, this yields 112 matched annotator--plot pairs. For each explanation, we first average its four criterion ratings to obtain an overall score and then compute the paired difference between \modelnamenc{} and the Base model within each matched pair.

\modelnamenc{} receives a higher overall score in 104 of 112 matched comparisons (92.9\%), while the Base model scores higher in 4 comparisons (3.6\%) and 4 are ties (3.6\%). Across the 112 matched pairs, the mean paired improvement is $+1.85$ points on the 5-point scale, with a 95\% confidence interval of $[1.66,\,2.05]$. These results provide human-evaluation evidence that the explanations produced by \modelnamenc{} are more visually grounded, better aligned with the annotated anomaly type, and less likely to contain unsupported claims than those of the Base model.

\noindent\textbf{Annotator instructions.} Annotators used a web interface that shows, for each item, the time-series plot with the ground-truth interval shaded red and the explanation's predicted interval shaded blue, the ground-truth interval, and one anonymized explanation. The instructions read: ``Rate the anonymized explanation from 1 to 5. Higher scores should mean better reasoning; for unsupported claims, 5 means the explanation avoids unsupported claims.'' The four questions are: (1) \emph{Are the explanation's claims actually visible in the plot?} (1 = not supported, 5 = clearly supported); (2) \emph{Does it correctly describe the anomaly type, such as spike, drop, shift, trend change, or variance change?} (1 = wrong characterization, 5 = correct characterization); (3) \emph{Does it avoid unsupported patterns, values, timestamps, or causes?} (1 = many unsupported claims, 5 = no unsupported claims); (4) \emph{Does the explanation provide sufficient details to support the final anomaly decision?} (1 = insufficient detail, 5 = sufficient detail). Annotators identified themselves only by a self-chosen anonymous code, no personal information was collected, and participation was voluntary and unpaid.

%% file: assets/appendix/statistical_support.tex
\begin{table*}[t]
\centering
\small
\caption{\textbf{Multi-run performance on \datasetnc{}.}
Results are reported as mean $\pm$ standard deviation across repeated runs.}
\label{tab:multirun_visanom}
\vspace{-0.2cm}
\resizebox{\textwidth}{!}{%
\begin{tabular}{lcccccc}
\toprule
\textbf{Model} &
\textbf{TP} &
\textbf{FP} &
\textbf{FN} &
\textbf{Precision (\%)} &
\textbf{Recall (\%)} &
\textbf{F\textsubscript{1} (\%)} \\
\midrule
TimesFM
& $196.00 \pm 0.00$
& $543.00 \pm 0.00$
& $563.00 \pm 0.00$
& $26.52 \pm 0.00$
& $25.82 \pm 0.00$
& $26.17 \pm 0.00$ \\

VLM4TS
& $437.80 \pm 5.07$
& $602.40 \pm 14.99$
& $321.20 \pm 5.07$
& $42.09 \pm 0.85$
& $57.68 \pm 0.67$
& $48.67 \pm 0.79$ \\

IForest
& $355.00 \pm 6.78$
& $385.00 \pm 6.78$
& $404.00 \pm 6.78$
& $47.97 \pm 0.92$
& $46.77 \pm 0.89$
& $47.36 \pm 0.90$ \\

\modelname{} (3B)
& $576.00 \pm 6.96$
& $208.40 \pm 19.83$
& $183.00 \pm 6.96$
& $\mathbf{73.46 \pm 2.05}$
& $75.89 \pm 0.92$
& $\mathbf{74.65 \pm 1.48}$ \\

\modelname{} (7B)
& $578.60 \pm 4.45$
& $221.00 \pm 9.14$
& $180.40 \pm 4.45$
& $72.37 \pm 0.88$
& $\mathbf{76.23 \pm 0.59}$
& $74.25 \pm 0.62$ \\
\bottomrule
\end{tabular}%
}
\end{table*}

\begin{table*}[t]
\centering
\small
\caption{\textbf{Multi-run performance on TSB-AD-U.}
Results are reported as mean $\pm$ standard deviation across repeated runs. ``Aff.'' denotes affiliation-based metrics.}
\label{tab:multirun_tsb}
\vspace{-0.2cm}
\resizebox{\textwidth}{!}{%
\begin{tabular}{lcccccc}
\toprule
\textbf{Model} &
\textbf{Precision (\%)} &
\textbf{Recall (\%)} &
\textbf{F\textsubscript{1} (\%)} &
\textbf{Aff. Precision (\%)} &
\textbf{Aff. Recall (\%)} &
\textbf{Aff. F\textsubscript{1} (\%)} \\
\midrule
TimesFM
& $32.32 \pm 0.00$
& $43.21 \pm 0.00$
& $32.19 \pm 0.00$
& $71.09 \pm 0.00$
& $57.23 \pm 0.00$
& $60.08 \pm 0.00$ \\

VLM4TS
& $28.44 \pm 1.13$
& $37.41 \pm 3.43$
& $19.48 \pm 1.63$
& $71.81 \pm 2.01$
& $51.16 \pm 2.06$
& $53.49 \pm 1.81$ \\

IForest
& $12.61 \pm 0.27$
& $18.16 \pm 0.33$
& $13.24 \pm 0.27$
& $50.59 \pm 0.48$
& $41.72 \pm 0.30$
& $42.85 \pm 0.23$ \\

\modelname{} (3B)
& $\mathbf{77.82 \pm 0.54}$
& $39.83 \pm 1.59$
& $48.91 \pm 1.39$
& $95.59 \pm 0.59$
& $71.68 \pm 0.79$
& $78.50 \pm 0.77$ \\

\modelname{} (7B)
& $76.34 \pm 0.93$
& $\mathbf{56.44 \pm 8.95}$
& $\mathbf{59.74 \pm 5.69}$
& $\mathbf{96.18 \pm 0.77}$
& $\mathbf{78.93 \pm 4.08}$
& $\mathbf{83.78 \pm 3.06}$ \\
\bottomrule
\end{tabular}%
}
\end{table*}

\subsection{Multi-Run Evaluation and Statistical Stability}
\label{apd:statistical_stability}

To assess run-to-run variability, we repeat the evaluation over five runs and report the mean and standard deviation for each metric. TimesFM is deterministic under pretrained inference, resulting in identical predictions across runs. Results on \datasetnc{} and TSB-AD-U are reported in Tables~\ref{tab:multirun_visanom} and~\ref{tab:multirun_tsb}, respectively. On \datasetnc{}, both \modelnamenc{} variants show low run-to-run variability and consistently outperform the baselines. 

The 3B and 7B models achieve mean F\textsubscript{1} scores of $74.65\pm1.48$ and $74.25\pm0.62$, respectively, compared with $48.67\pm0.79$ for VLM4TS and $47.36\pm0.90$ for IForest. Similar trends hold on TSB-AD-U, where both variants substantially outperform the baselines in F\textsubscript{1} and affiliation-based F\textsubscript{1}. Although the 7B model exhibits greater variability in recall on TSB-AD-U, its average performance remains substantially higher than the compared baselines. Overall, the multi-run results indicate that the observed performance gains are consistent across repeated evaluations.

%% file: assets/appendix/additional_results.tex
\subsection{Comparison with Baselines Trained on \datasetnc{}}

We also compare \modelnamenc{} with representative TSAD baselines, including recurrent, convolutional, autoencoder-based, and transformer-based models. Except for TimesFM, these semi-supervised methods are trained separately for each test time series using anomaly-free data. Since \datasetnc{} does not provide a separate anomaly-free training sequence for each test series, we use all points before the first ground-truth anomaly for training and evaluate the trained model on the entire test time series. In contrast, TimesFM is fine-tuned once on the \datasetnc{} training split using its official fine-tuning procedure to evaluate whether adapting a time-series foundation model narrows the performance gap. 

As shown in~\cref{tab:dl_tsad_baselines}, conventional deep-learning TSAD baselines obtain substantially lower interval-level precision, recall, and F\textsubscript{1} than \modelnamenc{}. Among these models, AE achieves the strongest F\textsubscript{1} score. Fine-tuning TimesFM on the \datasetnc{} training split improves its F\textsubscript{1} from 26.17\% to 34.76\%, but it remains substantially below \modelnamenc{} at 73.94\%. This comparison indicates that the performance gap persists even when the time-series foundation model is adapted using the benchmark training data.

\begin{table}[!t]
\centering
\small
\setlength{\tabcolsep}{10pt}
\caption{\textbf{Comparison with TSAD baselines trained on \datasetnc{}.}
TimesFM is fine-tuned on the \datasetnc{} training split using its official fine-tuning procedure.}
\label{tab:dl_tsad_baselines}
\resizebox{0.95\linewidth}{!}{%
\begin{tabular}{lcccccc}
\toprule
\textbf{Model} & \textbf{TP} & \textbf{FP} & \textbf{FN} &
\textbf{Precision (\%)} & \textbf{Recall (\%)} & \textbf{F\textsubscript{1} (\%)} \\
\midrule
CS-LSTM~\citep{zhangcontextual} & 175 & 556 & 584 & 23.94 & 23.06 & 23.49 \\
Anomaly-Transformer~\citep{xuanomaly} & 86 & 551 & 673 & 13.50 & 11.33 & 12.32 \\
Image-Embedding-CAE~\citep{garcia2022temporal} & 211 & 466 & 548 & 31.17 & 27.80 & 29.39 \\
AE~\citep{sakurada2014anomaly} & 297 & 439 & 462 & 40.35 & 39.13 & 39.73 \\
CNN~\citep{paparrizos2022tsb} & 237 & 481 & 522 & 33.01 & 31.23 & 32.09 \\
LSTM~\citep{paparrizos2022tsb} & 209 & 525 & 550 & 28.47 & 27.54 & 28.00 \\
\midrule
TimesFM & 196 & 543 & 563 & 26.52 & 25.82 & 26.17 \\
TimesFM (fine-tuned) & 260 & 477 & 499 & 35.28 & 34.26 & 34.76 \\
\midrule
\modelname{} (7B) & 576 & 223 & 183 & \textbf{72.09} & \textbf{75.89} & \textbf{73.94} \\
\bottomrule
\end{tabular}%
}
\end{table}

%% file: assets/appendix/plot_render_robustness.tex
\section{Index mapping and localization errors}\label{apd:plot}
Predicting exact anomaly indices from a rendered plot is challenging because the model must map visual locations back to the x-axis coordinate system, often using only sparsely displayed tick labels. We account for such localization errors by reporting interval-overlap metrics in addition to event-level detection metrics; overlap is sensitive to shifted, overly broad, and overly narrow predicted intervals. During \datasetnc{} construction, candidate responses are also evaluated for consistency with the ground-truth anomaly interval and for axis awareness, discouraging predictions with unsupported indices, timestamps, or value ranges.

Empirically, 98.0\% of the intervals predicted by \modelnamenc{} fall within the visible x-axis range, indicating that out-of-range coordinate predictions are uncommon. Without applying the 100-step extension to predicted intervals, \modelnamenc{} reduces the fraction of predicted intervals with no overlap with a ground-truth anomaly, compared with interval-only SFT, from 71.2\% to 43.8\%, corresponding to a 38.5\% relative reduction. Equivalently, the fraction of predictions with nonzero ground-truth overlap increases from 28.8\% to 56.2\%, a 27.4-point absolute improvement (95.1\% relative). The remaining localization errors are primarily associated with shifted start or end boundaries, interpolation between sparse x-axis ticks, and larger absolute index errors for long time series. These results indicate that the dominant limitation is boundary precision rather than unconstrained coordinate generation.

%% file: assets/appendix/efficiency.tex
\section{Training and Inference Efficiency}\label{apd:train_inf}
\noindent Model training and evaluation were performed on a high-performance computing resource using  4 NVIDIA H100 GPUs, each with 80 GB of memory. 

\textbf{Training efficiency.} We fine-tune Qwen2.5-VL-3B and Qwen2.5-VL-7B using supervised next-token prediction on the selected structured outputs. We use LoRA, with 95,178,752 trainable parameters, corresponding to \textbf{1.13\% of the backbone}. Vision encoder parameters are tuned via LoRA~\citep{hu2022lora} adapters, and only the targeted projection modules are updated: q\_proj, k\_proj, v\_proj, o\_proj, gate\_proj, up\_proj, and down\_proj. This makes \modelnamenc{} substantially smaller and more efficient than frontier VLM baselines while preserving strong anomaly localization performance.

\textbf{Inference efficiency.}  We report the inference efficiency of \modelnamenc{} compared with representative vision-language baselines.
\modelnamenc{} requires \textbf{16.5 $\pm$ 2.3 seconds per input series} on average, and its runtime is largely insensitive to sequence length because it operates on rendered plots rather than raw sequences.
In contrast, VLM4TS~\citep{he2025harnessing} exhibits substantially higher latency, taking up to \textbf{452.6 seconds per series}.
Prompting-based methods that rely on GPT-4o, including AnomLLM~\citep{zhoucan}, LLM-TSAD~\citep{parkdelving}, and LLMAD~\citep{liu2025large}, are faster on average (3.8 seconds) but show high variance, with worst-case latency up to 69.5 seconds.
They also cannot process sequences longer than 14K points.

%% file: assets/appendix/limitations.tex
\section{Limitations}\label{apd:limit}
\modelnamenc{} is currently limited to univariate time series, as it operates on single-channel plots and reasons over visual patterns within a single temporal signal.
Performance also depends on the quality of the plotted visualization: the plot must expose a clearly observable trend or deviation, while overly zoomed-in or zoomed-out views may obscure relevant anomaly patterns.
This limitation stems from reliance on visual reasoning rather than direct access to raw numerical values.
In future work, this constraint could be alleviated by automatically generating plots at multiple temporal resolutions and allowing a vision--language model to identify views in which meaningful trends and anomalies are most clearly expressed.

This reliance on visual evidence can also lead to weakly grounded or hallucinated explanations when the anomaly is visually ambiguous or compressed at the rendered scale, suggesting the need for more systematic evaluation of explanation faithfulness.
Finally, \modelnamenc{} is trained with supervised fine-tuning only.
Reinforcement learning or preference optimization may further improve interval localization and reasoning quality, but requires stable training signals for structured interval predictions.
We leave the design of such reward-based optimization methods to future work.

\section{Broader Impacts}\label{apd:impact}
This work contributes a benchmark and modeling framework for interpretable time-series anomaly detection by enabling joint anomaly localization and explanation from visual representations. By emphasizing explanation-grounded supervision, the proposed approach supports more transparent analysis of anomaly detection results, which may benefit research and practical evaluation in time-series monitoring tasks. The method operates on univariate time-series plots and depends on the quality of visualizations and large pre-trained vision–language models, which may limit applicability in certain settings. This work is intended as a research contribution, and its outputs should be interpreted with appropriate domain knowledge. We do not identify any immediate negative impacts.

%% file: checklist.tex
\section*{NeurIPS Paper Checklist}

\begin{enumerate}

\item {\bf Claims}
    \item[] Question: Do the main claims made in the abstract and introduction accurately reflect the paper's contributions and scope?
    \item[] Answer: \answerYes{}
    \item[] Justification: The paper's contributions, listed at the end of introduction section, are fully supported by the material presented in the remainder of the paper.
    \item[] Guidelines:
    \begin{itemize}
        \item The answer \answerNA{} means that the abstract and introduction do not include the claims made in the paper.
        \item The abstract and/or introduction should clearly state the claims made, including the contributions made in the paper and important assumptions and limitations. A \answerNo{} or \answerNA{} answer to this question will not be perceived well by the reviewers. 
        \item The claims made should match theoretical and experimental results, and reflect how much the results can be expected to generalize to other settings. 
        \item It is fine to include aspirational goals as motivation as long as it is clear that these goals are not attained by the paper. 
    \end{itemize}

\item {\bf Limitations}
    \item[] Question: Does the paper discuss the limitations of the work performed by the authors?
    \item[] Answer:  \answerYes{}  
    \item[] Justification: The limitations of this work are discussed in Appendix~\ref{apd:limit}. 
    \item[] Guidelines:
    \begin{itemize}
        \item The answer \answerNA{} means that the paper has no limitation while the answer \answerNo{} means that the paper has limitations, but those are not discussed in the paper. 
        \item The authors are encouraged to create a separate ``Limitations'' section in their paper.
        \item The paper should point out any strong assumptions and how robust the results are to violations of these assumptions (e.g., independence assumptions, noiseless settings, model well-specification, asymptotic approximations only holding locally). The authors should reflect on how these assumptions might be violated in practice and what the implications would be.
        \item The authors should reflect on the scope of the claims made, e.g., if the approach was only tested on a few datasets or with a few runs. In general, empirical results often depend on implicit assumptions, which should be articulated.
        \item The authors should reflect on the factors that influence the performance of the approach. For example, a facial recognition algorithm may perform poorly when image resolution is low or images are taken in low lighting. Or a speech-to-text system might not be used reliably to provide closed captions for online lectures because it fails to handle technical jargon.
        \item The authors should discuss the computational efficiency of the proposed algorithms and how they scale with dataset size.
        \item If applicable, the authors should discuss possible limitations of their approach to address problems of privacy and fairness.
        \item While the authors might fear that complete honesty about limitations might be used by reviewers as grounds for rejection, a worse outcome might be that reviewers discover limitations that aren't acknowledged in the paper. The authors should use their best judgment and recognize that individual actions in favor of transparency play an important role in developing norms that preserve the integrity of the community. Reviewers will be specifically instructed to not penalize honesty concerning limitations.
    \end{itemize}

\item {\bf Theory assumptions and proofs}
    \item[] Question: For each theoretical result, does the paper provide the full set of assumptions and a complete (and correct) proof?
    \item[] Answer: \answerNA{} 
    \item[] Justification: This work is empirical and does not include formal theoretical results or proofs.
    \item[] Guidelines:
    \begin{itemize}
        \item The answer \answerNA{} means that the paper does not include theoretical results. 
        \item All the theorems, formulas, and proofs in the paper should be numbered and cross-referenced.
        \item All assumptions should be clearly stated or referenced in the statement of any theorems.
        \item The proofs can either appear in the main paper or the supplemental material, but if they appear in the supplemental material, the authors are encouraged to provide a short proof sketch to provide intuition. 
        \item Inversely, any informal proof provided in the core of the paper should be complemented by formal proofs provided in appendix or supplemental material.
        \item Theorems and Lemmas that the proof relies upon should be properly referenced. 
    \end{itemize}

    \item {\bf Experimental result reproducibility}
    \item[] Question: Does the paper fully disclose all the information needed to reproduce the main experimental results of the paper to the extent that it affects the main claims and/or conclusions of the paper (regardless of whether the code and data are provided or not)?
    \item[] Answer: \answerYes{} 
    \item[] Justification: All settings, datasets, and prompts are detailed in \cref{sec:experiments} and Appendix \ref{apd:benchmark}.
    \item[] Guidelines:
    \begin{itemize}
        \item The answer \answerNA{} means that the paper does not include experiments.
        \item If the paper includes experiments, a \answerNo{} answer to this question will not be perceived well by the reviewers: Making the paper reproducible is important, regardless of whether the code and data are provided or not.
        \item If the contribution is a dataset and\slash or model, the authors should describe the steps taken to make their results reproducible or verifiable. 
        \item Depending on the contribution, reproducibility can be accomplished in various ways. For example, if the contribution is a novel architecture, describing the architecture fully might suffice, or if the contribution is a specific model and empirical evaluation, it may be necessary to either make it possible for others to replicate the model with the same dataset, or provide access to the model. In general. releasing code and data is often one good way to accomplish this, but reproducibility can also be provided via detailed instructions for how to replicate the results, access to a hosted model (e.g., in the case of a large language model), releasing of a model checkpoint, or other means that are appropriate to the research performed.
        \item While NeurIPS does not require releasing code, the conference does require all submissions to provide some reasonable avenue for reproducibility, which may depend on the nature of the contribution. For example
        \begin{enumerate}
            \item If the contribution is primarily a new algorithm, the paper should make it clear how to reproduce that algorithm.
            \item If the contribution is primarily a new model architecture, the paper should describe the architecture clearly and fully.
            \item If the contribution is a new model (e.g., a large language model), then there should either be a way to access this model for reproducing the results or a way to reproduce the model (e.g., with an open-source dataset or instructions for how to construct the dataset).
            \item We recognize that reproducibility may be tricky in some cases, in which case authors are welcome to describe the particular way they provide for reproducibility. In the case of closed-source models, it may be that access to the model is limited in some way (e.g., to registered users), but it should be possible for other researchers to have some path to reproducing or verifying the results.
        \end{enumerate}
    \end{itemize}

\item {\bf Open access to data and code}
    \item[] Question: Does the paper provide open access to the data and code, with sufficient instructions to faithfully reproduce the main experimental results, as described in supplemental material?
    \item[] Answer: \answerNo{} 
    \item[] Justification: The code, benchmark data, and reproduction instructions will be made publicly available by the time of publication.
    \item[] Guidelines:
    \begin{itemize}
        \item The answer \answerNA{} means that paper does not include experiments requiring code.
        \item Please see the NeurIPS code and data submission guidelines (\url{https://neurips.cc/public/guides/CodeSubmissionPolicy}) for more details.
        \item While we encourage the release of code and data, we understand that this might not be possible, so \answerNo{} is an acceptable answer. Papers cannot be rejected simply for not including code, unless this is central to the contribution (e.g., for a new open-source benchmark).
        \item The instructions should contain the exact command and environment needed to run to reproduce the results. See the NeurIPS code and data submission guidelines (\url{https://neurips.cc/public/guides/CodeSubmissionPolicy}) for more details.
        \item The authors should provide instructions on data access and preparation, including how to access the raw data, preprocessed data, intermediate data, and generated data, etc.
        \item The authors should provide scripts to reproduce all experimental results for the new proposed method and baselines. If only a subset of experiments are reproducible, they should state which ones are omitted from the script and why.
        \item At submission time, to preserve anonymity, the authors should release anonymized versions (if applicable).
        \item Providing as much information as possible in supplemental material (appended to the paper) is recommended, but including URLs to data and code is permitted.
    \end{itemize}

\item {\bf Experimental setting/details}
    \item[] Question: Does the paper specify all the training and test details (e.g., data splits, hyperparameters, how they were chosen, type of optimizer) necessary to understand the results?
    \item[] Answer: \answerYes{} 
    \item[] Justification:  Data splits, prompts, LLM versions, and ablation setups are described in \cref{sec:experiments}, \cref{sec:ablation},  Appendix~\ref{apd:benchmark} and Appendix~\ref{apd:imple}.
    \item[] Guidelines:
    \begin{itemize}
        \item The answer \answerNA{} means that the paper does not include experiments.
        \item The experimental setting should be presented in the core of the paper to a level of detail that is necessary to appreciate the results and make sense of them.
        \item The full details can be provided either with the code, in appendix, or as supplemental material.
    \end{itemize}

\item {\bf Experiment statistical significance}
    \item[] Question: Does the paper report error bars suitably and correctly defined or other appropriate information about the statistical significance of the experiments?
    \item[] Answer: \answerYes{} 
    \item[] Justification: Appendix~\ref{apd:statistical_stability} reports the mean and standard deviation over five runs on \datasetnc{} and \textsc{TSB-AD-U} (Tables~\ref{tab:multirun_visanom} and~\ref{tab:multirun_tsb}), and Appendix~\ref{apd:faithfulness} reports a 95\% confidence interval for the paired improvement in the human evaluation.
 
    \item[] Guidelines:
    \begin{itemize}
        \item The answer \answerNA{} means that the paper does not include experiments.
        \item The authors should answer \answerYes{} if the results are accompanied by error bars, confidence intervals, or statistical significance tests, at least for the experiments that support the main claims of the paper.
        \item The factors of variability that the error bars are capturing should be clearly stated (for example, train/test split, initialization, random drawing of some parameter, or overall run with given experimental conditions).
        \item The method for calculating the error bars should be explained (closed form formula, call to a library function, bootstrap, etc.)
        \item The assumptions made should be given (e.g., Normally distributed errors).
        \item It should be clear whether the error bar is the standard deviation or the standard error of the mean.
        \item It is OK to report 1-sigma error bars, but one should state it. The authors should preferably report a 2-sigma error bar than state that they have a 96\% CI, if the hypothesis of Normality of errors is not verified.
        \item For asymmetric distributions, the authors should be careful not to show in tables or figures symmetric error bars that would yield results that are out of range (e.g., negative error rates).
        \item If error bars are reported in tables or plots, the authors should explain in the text how they were calculated and reference the corresponding figures or tables in the text.
    \end{itemize}

\item {\bf Experiments compute resources}
    \item[] Question: For each experiment, does the paper provide sufficient information on the computer resources (type of compute workers, memory, time of execution) needed to reproduce the experiments?
    \item[] Answer: \answerYes{} 
    \item[] Justification: Details on computational resources, training efficiency, and inference efficiency are included in Appendix~\ref{apd:train_inf}.
    \item[] Guidelines:
    \begin{itemize}
        \item The answer \answerNA{} means that the paper does not include experiments.
        \item The paper should indicate the type of compute workers CPU or GPU, internal cluster, or cloud provider, including relevant memory and storage.
        \item The paper should provide the amount of compute required for each of the individual experimental runs as well as estimate the total compute. 
        \item The paper should disclose whether the full research project required more compute than the experiments reported in the paper (e.g., preliminary or failed experiments that didn't make it into the paper). 
    \end{itemize}
    
\item {\bf Code of ethics}
    \item[] Question: Does the research conducted in the paper conform, in every respect, with the NeurIPS Code of Ethics \url{https://neurips.cc/public/EthicsGuidelines}?
    \item[] Answer: \answerYes{} 
    \item[] Justification:  The research complies with NeurIPS ethical guidelines.
    \item[] Guidelines:
    \begin{itemize}
        \item The answer \answerNA{} means that the authors have not reviewed the NeurIPS Code of Ethics.
        \item If the authors answer \answerNo, they should explain the special circumstances that require a deviation from the Code of Ethics.
        \item The authors should make sure to preserve anonymity (e.g., if there is a special consideration due to laws or regulations in their jurisdiction).
    \end{itemize}

\item {\bf Broader impacts}
    \item[] Question: Does the paper discuss both potential positive societal impacts and negative societal impacts of the work performed?
    \item[] Answer: \answerYes{}  
    \item[] Justification: We discuss broader impacts in Appendix~\ref{apd:impact}.
    \item[] Guidelines:
    \begin{itemize}
        \item The answer \answerNA{} means that there is no societal impact of the work performed.
        \item If the authors answer \answerNA{} or \answerNo, they should explain why their work has no societal impact or why the paper does not address societal impact.
        \item Examples of negative societal impacts include potential malicious or unintended uses (e.g., disinformation, generating fake profiles, surveillance), fairness considerations (e.g., deployment of technologies that could make decisions that unfairly impact specific groups), privacy considerations, and security considerations.
        \item The conference expects that many papers will be foundational research and not tied to particular applications, let alone deployments. However, if there is a direct path to any negative applications, the authors should point it out. For example, it is legitimate to point out that an improvement in the quality of generative models could be used to generate Deepfakes for disinformation. On the other hand, it is not needed to point out that a generic algorithm for optimizing neural networks could enable people to train models that generate Deepfakes faster.
        \item The authors should consider possible harms that could arise when the technology is being used as intended and functioning correctly, harms that could arise when the technology is being used as intended but gives incorrect results, and harms following from (intentional or unintentional) misuse of the technology.
        \item If there are negative societal impacts, the authors could also discuss possible mitigation strategies (e.g., gated release of models, providing defenses in addition to attacks, mechanisms for monitoring misuse, mechanisms to monitor how a system learns from feedback over time, improving the efficiency and accessibility of ML).
    \end{itemize}
    
\item {\bf Safeguards}
    \item[] Question: Does the paper describe safeguards that have been put in place for responsible release of data or models that have a high risk for misuse (e.g., pre-trained language models, image generators, or scraped datasets)?
    \item[] Answer:  \answerNA{}
    \item[] Justification: The paper does not release any high-risk models or datasets.
    \item[] Guidelines:
    \begin{itemize}
        \item The answer \answerNA{} means that the paper poses no such risks.
        \item Released models that have a high risk for misuse or dual-use should be released with necessary safeguards to allow for controlled use of the model, for example by requiring that users adhere to usage guidelines or restrictions to access the model or implementing safety filters. 
        \item Datasets that have been scraped from the Internet could pose safety risks. The authors should describe how they avoided releasing unsafe images.
        \item We recognize that providing effective safeguards is challenging, and many papers do not require this, but we encourage authors to take this into account and make a best faith effort.
    \end{itemize}

\item {\bf Licenses for existing assets}
    \item[] Question: Are the creators or original owners of assets (e.g., code, data, models), used in the paper, properly credited and are the license and terms of use explicitly mentioned and properly respected?
    \item[] Answer: \answerYes{} 
    \item[] Justification: All assets are properly cited.
    \item[] Guidelines:
    \begin{itemize}
        \item The answer \answerNA{} means that the paper does not use existing assets.
        \item The authors should cite the original paper that produced the code package or dataset.
        \item The authors should state which version of the asset is used and, if possible, include a URL.
        \item The name of the license (e.g., CC-BY 4.0) should be included for each asset.
        \item For scraped data from a particular source (e.g., website), the copyright and terms of service of that source should be provided.
        \item If assets are released, the license, copyright information, and terms of use in the package should be provided. For popular datasets, \url{paperswithcode.com/datasets} has curated licenses for some datasets. Their licensing guide can help determine the license of a dataset.
        \item For existing datasets that are re-packaged, both the original license and the license of the derived asset (if it has changed) should be provided.
        \item If this information is not available online, the authors are encouraged to reach out to the asset's creators.
    \end{itemize}

\item {\bf New assets}
    \item[] Question: Are new assets introduced in the paper well documented and is the documentation provided alongside the assets?
    \item[] Answer: \answerYes{} 
    \item[] Justification: The datasets and baselines used in this work are publicly available with accompanying documentation. Details of the proposed model, benchmark construction, implementation, training setup, and ablation studies are provided in \cref{sec:experiments}, \cref{sec:ablation}, Appendix~\ref{apd:benchmark} and Appendix~\ref{apd:imple}. The model checkpoints and associated documentation will be made publicly available by the time of publication.
    \item[] Guidelines:
    \begin{itemize}
        \item The answer \answerNA{} means that the paper does not release new assets.
        \item Researchers should communicate the details of the dataset\slash code\slash model as part of their submissions via structured templates. This includes details about training, license, limitations, etc. 
        \item The paper should discuss whether and how consent was obtained from people whose asset is used.
        \item At submission time, remember to anonymize your assets (if applicable). You can either create an anonymized URL or include an anonymized zip file.
    \end{itemize}

\item {\bf Crowdsourcing and research with human subjects}
    \item[] Question: For crowdsourcing experiments and research with human subjects, does the paper include the full text of instructions given to participants and screenshots, if applicable, as well as details about compensation (if any)? 
    \item[] Answer: \answerYes{} 
    \item[] Justification: Appendix~\ref{apd:faithfulness} gives the full instructions and rating questions shown to the six annotators and describes the study interface. No crowdsourcing platform was used, and annotators were unpaid volunteers.

    \item[] Guidelines:
    \begin{itemize}
        \item The answer \answerNA{} means that the paper does not involve crowdsourcing nor research with human subjects.
        \item Including this information in the supplemental material is fine, but if the main contribution of the paper involves human subjects, then as much detail as possible should be included in the main paper. 
        \item According to the NeurIPS Code of Ethics, workers involved in data collection, curation, or other labor should be paid at least the minimum wage in the country of the data collector. 
    \end{itemize}

\item {\bf Institutional review board (IRB) approvals or equivalent for research with human subjects}
    \item[] Question: Does the paper describe potential risks incurred by study participants, whether such risks were disclosed to the subjects, and whether Institutional Review Board (IRB) approvals (or an equivalent approval/review based on the requirements of your country or institution) were obtained?
    \item[] Answer: \answerNo{} 
    \item[] Justification: No IRB review was sought. The human evaluation posed minimal risk: volunteer annotators only rated model-generated explanations of time-series plots, identified themselves by a self-chosen anonymous code, and no personal or sensitive information was collected (Appendix~\ref{apd:faithfulness}).
    \item[] Guidelines:
    \begin{itemize}
        \item The answer \answerNA{} means that the paper does not involve crowdsourcing nor research with human subjects.
        \item Depending on the country in which research is conducted, IRB approval (or equivalent) may be required for any human subjects research. If you obtained IRB approval, you should clearly state this in the paper. 
        \item We recognize that the procedures for this may vary significantly between institutions and locations, and we expect authors to adhere to the NeurIPS Code of Ethics and the guidelines for their institution. 
        \item For initial submissions, do not include any information that would break anonymity (if applicable), such as the institution conducting the review.
    \end{itemize}

\item {\bf Declaration of LLM usage}
    \item[] Question: Does the paper describe the usage of LLMs if it is an important, original, or non-standard component of the core methods in this research? Note that if the LLM is used only for writing, editing, or formatting purposes and does \emph{not} impact the core methodology, scientific rigor, or originality of the research, declaration is not required.
    \item[] Answer: \answerYes{} 
    \item[] Justification: The proposed method is built upon pretrained vision--language models (Qwen2.5-VL-3B and Qwen2.5-VL-7B), which are supervised fine-tuned for time-series anomaly localization and reasoning generation. Large vision--language models are also used during dataset construction to generate candidate explanations and structured outputs for supervision selection. The LLM usage is fully described in the paper.
    \item[] Guidelines:
    \begin{itemize}
        \item The answer \answerNA{} means that the core method development in this research does not involve LLMs as any important, original, or non-standard components.
        \item Please refer to our LLM policy in the NeurIPS handbook for what should or should not be described.
    \end{itemize}

\end{enumerate}